%% file: main.tex
\documentclass[runningheads]{llncs}

\usepackage{eccvabbrv}
\usepackage{lineno}

\usepackage{graphicx}
\usepackage{booktabs}

\usepackage{natbib}
\usepackage{wrapfig}
\usepackage{multirow}
\usepackage[svgnames]{xcolor}
\usepackage{nicematrix}
\usepackage{makecell}
\usepackage[accsupp]{axessibility}  

\usepackage{hyperref}
\usepackage{setspace}

\usepackage{orcidlink}

\usepackage[table]{xcolor}
\usepackage{booktabs}

\definecolor{inductivehead}{RGB}{255,238,221}
\definecolor{discoveryhead}{RGB}{229,245,224}
\definecolor{predictionhead}{RGB}{239,237,245}

\definecolor{inductivehead}{RGB}{245,160,105}
\definecolor{discoveryhead}{RGB}{140,205,160}
\definecolor{predictionhead}{RGB}{190,180,225}
\definecolor{hybridhead}{RGB}{220,220,220}

\begin{document}


\title{Finding Needles in the Haystack: Transductive Active Labeling in Ecology} 

\author{Rupa Kurinchi-Vendhan\inst{1} \and
Sara Beery\inst{1}} 
\institute{Massachusetts Institute of Technology, Cambridge MA 02139, USA\\ Correspondence to: \email{rupak272@mit.edu}}

\maketitle

\begin{abstract}
    Active learning is now standard practice in labeling ecological data, enabling ecologists to quickly process large volumes of field data to understand and monitor natural environments. Current practices evaluate active learning \textit{inductively}, estimating predictive performance on a held-out test set. We argue that this evaluation is misaligned with most ecological tasks, where the goal is to \textit{transductively} label an entire pool of data as efficiently as possible. We demonstrate that ignoring the human-in-the-loop underestimates the importance of continuing to label, particularly for classes in the long tail which may be of disproportionate ecological importance (rare species, uncommon behaviors, etc.). Our analysis shows that, for this long tail, the transductive objective shifts importance from prediction to \textit{discovery}: the true challenge becomes finding ``needles in the haystack,'' examples of rare classes that are embedded within dense regions of abundant classes in the latent geometry, which we quantify with a novel metric of \textit{sampling difficulty}. Finally, to translate these insights to practical ecological workflows, we propose a conservative hybrid stopping criterion inspired by ecological rarefaction curves, and show that combining predictive performance with discovery criteria reduces premature stopping on long-tailed pools, improving rare-class recovery when discovery—not classification—is the limiting factor. 
    
    \textbf{Data/Code Availability:} All data preprocessing and analysis scripts available \href{https://anonymous.4open.science/r/Transductive-AL-4D9F/README.md}{here}. This codebase will become publicly available upon publication.

\end{abstract}

\section{Introduction}
\label{sec:intro}

Ecologists and conservation scientists are now collecting environmental data at a scale that far exceeds what experts can manually annotate. A single passive acoustic recorder running continuously produces over 8{,}760 hours of audio per year, and monitoring campaigns often deploy dozens of such sensors across a study area \cite{sugai2019terrestrial}. Autonomous camera traps have given rise to programs such as Snapshot Safari, which spans 15 imaging projects across 1{,}824 camera locations \cite{pardo2021snapshot}. Across modalities, the same core bottleneck persists: manual annotation by domain experts cannot keep pace with data collection.

\begin{figure}[h]
    \centering
    \includegraphics[width=\linewidth]{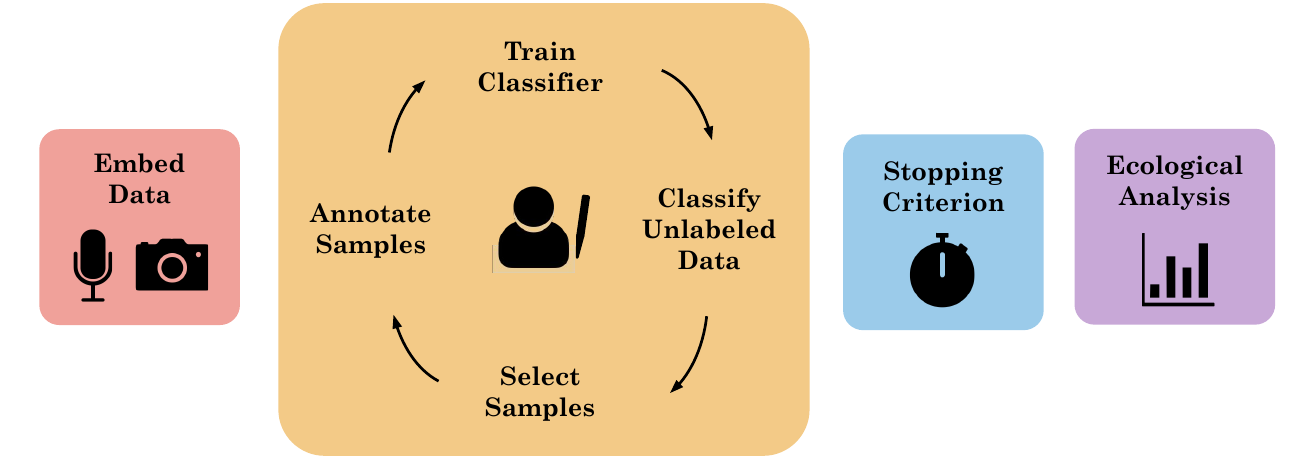}
    \caption{\textbf{Framework for active labeling used in practice for ecological data.}} 
    \label{fig:setting}
    \vspace{-5mm}
\end{figure}

Today, AI-enabled ecological workflows often begin with a fixed deployment pool, embed each sample using a pretrained encoder, and train a lightweight task-specific model over these representations. Active labeling iteratively selects samples for expert annotation, updates the model, and propagates predictions across the remaining pool. This paradigm is common in bioacoustics \cite{ghani2023global,van2025perch,burns2025perch,hagiwara2023beans,dumoulin2025search} and is increasingly being extended to camera-trap imagery and remote sensing, where pretrained representations support large-scale ecological classification and monitoring \cite{markoff2026vision,fabian2023multimodal,dussert2025zero,burges2025active}. These pipelines rely on simple updates, such as linear probes on frozen embeddings. In these settings, label efficiency depends primarily on which examples are selected, rather than on model retraining. Figure~\ref{fig:setting} illustrates the active-labeling framework studied here.

Most active labeling benchmarks in ecology use an \textbf{\textit{inductive evaluation}} framework, measuring predictive performance on a held-out, labeled test set \citep{ghani2023global,miron2026avex,van2025perch,hamer2023birb, rauch2024birdset, hagiwara2023beans}. Evaluation in these settings is still largely summarized through predictive metrics on annotated test splits as a function of labeling budget \citep{zhan2021comparative, meduri2020comprehensive, settles2009active}. 

\begin{figure}[h]
    \centering
    \includegraphics[width=\linewidth]{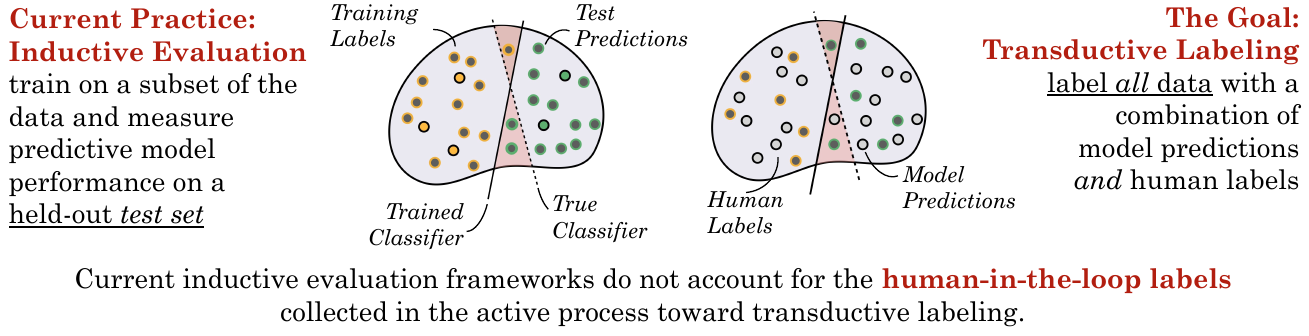}
    \caption{\textbf{Held-out test performance does not capture the \textit{transductive} objective of labeling an entire pool using both human and model-generated labels.} In this work, we show that accounting for human-provided labels substantially changes measured performance, especially for rare categories.} 
    \label{fig:setting2}
\end{figure}

In practice, ecologists are often less concerned with training a classifier for future observations than with assigning labels to a fixed set of observations collected during a particular survey, season, or location. Researchers aim to identify which taxa are present, surface rare or previously unobserved events, and estimate ecological quantities of interest within the dataset itself \cite{aide2013real,sueur2008rapid,gibb2019emerging,norouzzadeh2018automatically,steenweg2017scaling,swanson2015snapshot}. We define this setting as \textbf{\textit{transductive active labeling}}, where the objective is to accurately label a fixed pool of data with minimal human input. Here, performance is calculated on the {entire pool}, reflecting \textit{both} classifier predictions on the remaining unlabeled data and the direct correction provided by human annotations as samples are labeled.  

Most inductive benchmarks do not explicitly capture the true labeling progress with the iteratively growing set of human-in-the-loop labels from active learning, and may overlook the factors that determine labeling efficiency in practice. This distinction is illustrated in Figure \ref{fig:setting2}. In this work, 
\noindent \begin{enumerate}
    \item we argue that inductive evaluation is misaligned with transductive objectives, and show empirically that the core challenge shifts from classification to \textit{discovery} for the long tail;
    \item we introduce a formal metric of \emph{discoverability} that describes the difficulty of sampling data, and further formalize \emph{needles in haystacks} as data that are highly difficult to sample;
    \item we propose a novel \textit{stopping criterion} for labeling based on both performance and discovery rates, inspired by ecological rarefaction curves, and show that it leads to better performance across rare categories than purely predictive performance-based criteria.
\end{enumerate}

\section{Methods}
\label{sec:methods}

\vspace{5pt}\noindent \textbf{Formal Definitions.}
We study the standard pool-based active learning setting, adapted to the transductive objective of fixed-pool ecological labeling. Let $\mathcal{D} = \{x_i\}_{i=1}^N$ denote a fixed pool of unlabeled data that an expert wishes to annotate. Active learning proceeds in cycles, with a fixed batch size of 10 samples per iteration; in Appendix Figure~\ref{fig:budget}, we show that varying the batch size has little effect. The first batch is selected uniformly at random to initialize the labeled set. At iteration $t$, the labeled set is $\mathcal{L}_t \subset \mathcal{D}$ and the remaining unlabeled pool is $\mathcal{U}_t = \mathcal{D} \setminus \mathcal{L}_t$. A query function $q(\cdot)$ selects a batch of samples from $\mathcal{U}_t$ for expert annotation, after which the newly labeled samples are added to $\mathcal{L}_t$.

Recent ecological active learning systems increasingly operate on pretrained feature embeddings, especially in bioacoustics, where large pretrained encoders such as Perch, BirdNET, and related models are used as general-purpose representation extractors \citep{ghani2023global,van2025perch,kahl2021birdnet,dumoulin2025search}. Many practical workflows then train a lightweight downstream classifier, such as a linear probe, over frozen embeddings. In this setting, label efficiency is determined largely by which examples are selected for annotation rather than by end-to-end representation learning. We therefore isolate the sampling component by fixing the embedding model and retraining a multinomial logistic-regression probe from scratch at each active learning cycle. The probe operates on frozen embeddings and uses the resulting class probabilities for uncertainty-based query strategies; we compare linear probes and MLPs, with complete implementation details provided in Appendix~\ref{app:classifier_details}.

\paragraph{Inductive evaluation.}
Most active learning pipelines are evaluated in the inductive setting, where predictive performance is measured on a held-out test set, $\mathcal{D}_{\text{test}} \subset \mathcal{D}$. This test set is disjoint from both the labeled set $\mathcal{L}_t$ and the unlabeled pool $\mathcal{U}_t$, and is never accessed during sample selection or training. For each dataset, we reserve 20\% of examples for inductive evaluation, while the remaining 80\% constitutes the fixed pool used for transductive active labeling. We do not use a separate validation split. 
After each acquisition step, the classifier is trained on $\mathcal{L}_t$ and evaluated on $\mathcal{D}_{\text{test}}$. Under this protocol, labeling decisions affect performance only indirectly, through improvements in the classifier’s ability to generalize to unseen examples. We use the default splits provided by each dataset source; when none are available, we create a random 20\% held-out split.

\paragraph{Transductive objective.}
We adopt a transductive labeling perspective, where performance is considered over the full deployment pool $\mathcal{D}$ rather than on a separate held-out test set. At step $t$, examples in $\mathcal{L}_t$ are considered ``correct'' because they have been labeled by experts, while performance on examples in $\mathcal{U}_t$ depends on the predictions of the classifier $f_t$ trained on the labeled data. Therefore, overall performance reflects both the examples already handled by expert annotation and the classifier’s accuracy on the remaining unlabeled portion of the pool.

This is not intended as a practical evaluation procedure, since measuring full-pool accuracy would require ground-truth labels for all examples. Instead, it provides a \textit{post-hoc diagnostic} to reason about actual progress during active labeling. The distinction from inductive evaluation is important: under inductive evaluation, performance is measured on a disjoint test set, so labeling decisions matter only indirectly through improved generalization. Under the transductive labeling perspective, labeling decisions also improve realized system performance directly, because every newly annotated example is immediately resolved. As we show below, this pool-level perspective can provide a more appropriate basis for deciding when additional expert labeling is no longer warranted.

\vspace{5pt}\noindent \textbf{Datasets.}
In this paper, we study bioacoustic and camera-trap datasets, where annotation is often costly, expert-driven, and conducted at scale. We evaluate on the classification subset of the BEANS benchmark \citep{hagiwara2023beans}, together with ReefSet \citep{williams2025using} and camera-trap datasets from the Snapshot Safari network \citep{pardo2021snapshot}, as summarized in Table~\ref{tab:datasets}.

\begin{table}[h]
\centering
\caption{\textbf{Datasets used in our analysis.}}
\label{tab:datasets}
\scriptsize
\begin{tabular}{p{7.2cm}cc}
\toprule
\textbf{Dataset} & \textbf{No. Samples} & \textbf{No. Classes} \\
\toprule
Watkins Marine Mammals \citep{sayigh2016watkins} \newline
\textit{whales, dolphins, seals, and walruses (44.1 kHz)}
& 1,695 & 32 \\
\hline
Cornell Birdcall Identification (CBI) \citep{cornell2020birdcall} \newline
\textit{birds (44.1 kHz)}
& 21,375 & 264 \\
\hline
HumBugDB \citep{kiskin2021humbugdb} \newline
\textit{mosquito wingbeats (44.1 kHz)}
& 13,011 & 14 \\
\hline
ReefSet \citep{williams2025using} \newline
\textit{reef biophony, anthrophony, and geophony (16 kHz)}
& 57,074 & 37 \\
\hline
Egyptian Fruit Bats (EFB) \citep{prat2017annotated} \newline
\textit{individual bats (250 kHz)}
& 10,000 & 10 \\
\hline
Dogs \citep{yin2004barking} \newline
\textit{individual domestic dogs (44.1 kHz)}
& 693 & 10 \\
\toprule
Snapshot Kruger \citep{pardo2021snapshot} \newline
\textit{camera-trap data from Kruger National Park}
& 10,072 & 46 \\
\hline
Snapshot Camdeboo \citep{pardo2021snapshot} \newline
\textit{camera-trap data from Camdeboo National Park}
& 30,227 & 43 \\
\hline
Snapshot Kgalagadi \citep{pardo2021snapshot} \newline
\textit{camera-trap data from the Kgalagadi Transfrontier Park}
& 10,222 & 31 \\
\bottomrule
\end{tabular}
\end{table}

 \vspace{5pt}\noindent \textbf{Embedding Models.}
For bioacoustic data, we evaluate several recent foundation models for representation extraction. We consider Perch 2.0 \citep{van2025perch}, Perch 1.0 \citep{ghani2023global}, SurfPerch \citep{williams2025using}, BirdNET \citep{kahl2021birdnet}, and BEATS \citep{chen2022beats}. Unless specified, all experiments below use Perch 2.0 embeddings for bioacoustic datasets, as they perform best across datasets \citep{van2025perch}; a comparison to other embedding models is provided in the Appendix \ref{app:embedding_geometry}. For image data, we use DINOv3 embeddings \citep{simeoni2025dinov3}, which provide strong general-purpose visual representations and have shown strong zero-shot performance on animal images \citep{markoff2026vision}. 

\vspace{5pt}\noindent \textbf{Sampling Strategies.}
Active learning aims to reduce labeling cost by iteratively selecting the most informative examples for annotation. Early works introduce pool-based active learning, where a model queries labels from a fixed unlabeled pool and is retrained after each acquisition step \citep{lewis1995sequential}. Research in this area primarily focuses on sampling strategies designed to maximize information gain  \citep{weinstein2019selective,settles2009active,fu2013survey}.

In this paper, we evaluate several commonly-used active learning query strategies \citep{settles2009active}: random sampling; least confidence, which prioritizes examples with the lowest maximum predicted class probability; margin sampling, which uses the smallest gap between the top two predicted class probabilities; entropy sampling, which uses the highest predictive entropy; BADGE, a gradient-based strategy combining uncertainty and diversity \citep{ash2019deep}; and coreset sampling, which maximizes coverage of the embedding space using a $k$-center objective \citep{sener2017active}.

\vspace{5pt}\noindent \textbf{Metrics.} Unless otherwise noted, all reported accuracy values reflect class-averaged classification accuracies, to account for severe class imbalance in ecological datasets.

\section{Results and Discussion}
\label{sec:results}

\vspace{5pt}\noindent \textbf{Comparing estimated inductive performance to actualized transductive performance.} For each active learning strategy, we measure performance using a \emph{normalized class-averaged accuracy} metric that subtracts the random-sampling baseline at each labeling budget. We define
\begin{equation}
    A_{\mathrm{norm}}(b) =
    \frac{A(b) - A_{\mathrm{rand}}(b)}
    {A_{\max}(b) - A_{\mathrm{rand}}(b)},
\label{eq:normalized_acc}
\end{equation}
where $A(b)$ denotes class-averaged accuracy after labeling $b$ samples, $A_{\mathrm{rand}}(b)$ is the corresponding class-averaged accuracy achieved by random sampling at the same budget, and $A_{\max}(b)$ is the empirical performance ceiling at budget $b$, defined as the maximum class-averaged accuracy achieved across all evaluated methods for that dataset. This normalization makes performance more comparable across datasets. A value of $A_{\mathrm{norm}}(b)=0$ indicates performance equal to random sampling, while $A_{\mathrm{norm}}(b)=1$ indicates performance at the empirical ceiling. We report $A_{\mathrm{norm}}(b)$ across labeling budgets and average across datasets for aggregate comparisons.

\begin{figure}[h]
    \centering
    \centering
    \includegraphics[width=\linewidth]{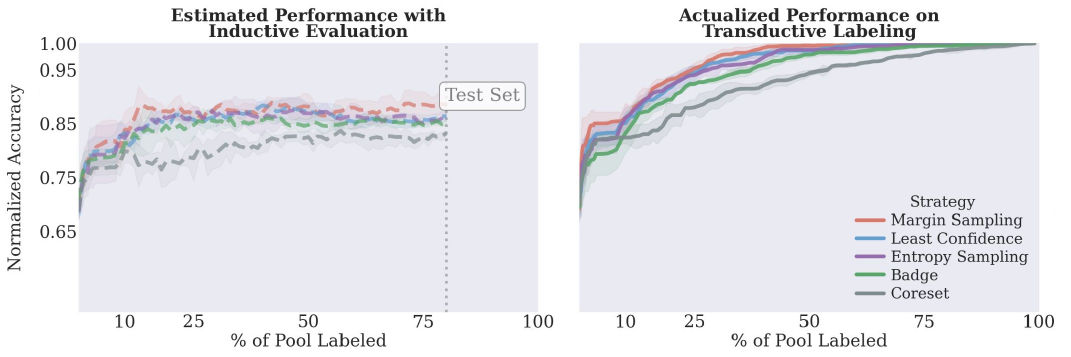}
    \vspace{-10pt}
    \caption{\textbf{Transductive performance continues to improve as more samples are labeled, while inductive evaluation is limited by test-set generalization.} Normalized accuracy, averaged across datasets, shows that inductive evaluation systematically underestimates the utility of active learning for fixed-pool ecological labeling.}
    \label{fig:inductive_vs_transductive}
    \vspace{-5mm}
\end{figure}

Figure~\ref{fig:inductive_vs_transductive} shows that inductive evaluation systematically misrepresents the actualized performance of active learning for fixed-pool deployments. Actualized transductive labeling performance is intuitively higher than estimated performance via inductive evaluation: once a sample is labeled, it directly contributes to the number of correctly labeled examples in the pool. Inductive evaluation does not account for this direct impact of having a human-in-the-loop, and thus only captures the indirect impact of improved model predictions when training additional human labels.

\begin{figure}[h]
    \centering
    \includegraphics[width=\linewidth]{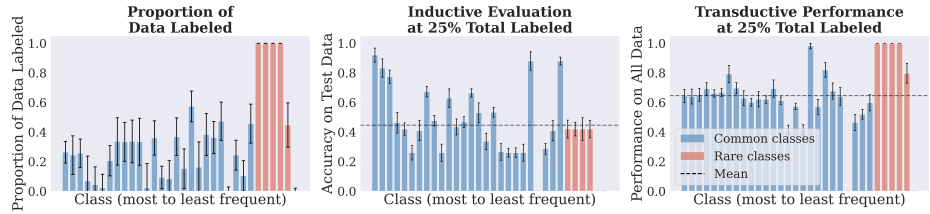}
    \caption{\textbf{Sampling and annotating, not classification, drives rare-category performance.} On the Watkins Marine Mammal dataset, inductive evaluation underestimates long-tail performance. After labeling $25\%$ of the dataset with margin sampling (left), predicted accuracy on the rarest 20\% of classes remains low on a held-out test set (middle). Once these classes are discovered in the fixed pool, they are labeled with high accuracy (right). Thus, rare-class outcomes are primarily limited by sampling, not classification.}
    \label{fig:per_class_sampling}
    \vspace{-5mm}
\end{figure}

\vspace{5pt}\noindent \textbf{Discovery, Not Classification Drives This Performance Gap.}
The gap between estimated and actual labeling performance stems from how rare classes skew the class-average accuracy. Figure~\ref{fig:per_class_sampling} illustrates this effect on the Watkins Marine Mammal dataset. Under inductive evaluation (middle), rare classes have low and unstable held-out accuracy. With limited labeled examples, the model cannot reliably predict categories it has barely observed. In contrast, under the transductive setting (right), accuracy within the fixed deployment pool is substantially higher, particularly for rare classes, as it accounts for correct human labeling when examples of those classes are sampled. The same pattern holds across datasets and acquisition strategies, as shown in the Appendix Figures \ref{fig:watkins_strat}, and \ref{fig:datasets_sampling_1}, \ref{fig:datasets_sampling_2}.

\textit{This observation reframes how we interpret rare class performance in learned systems}---prior work consistently documents poor performance for these categories \citep{settles2009active,kath2024leveraging}. In long-tailed ecological datasets, rare classes are often of disproportionate scientific interest (e.g. endangered species), but difficult to learn because they are infrequently observed during training. However, our results emphasize that, in a transductive setting, performance is fundamentally \emph{sampling-limited}, not classifier-limited. Thus, a system that predicts common classes well and surfaces rare classes for an expert to directly label would be both efficient and accurate. The primary bottleneck for rare-class performance in transductive active labeling is therefore \textit{discovery}: how efficiently these rare samples are surfaced.

\vspace{5pt}\noindent \textbf{What makes data systematically harder to discover?} Because acquisition functions operate on embeddings $z=f_\theta(x)$ without access to ground-truth labels at query time, selection decisions are governed by representation geometry. If long-tail performance is sampling-limited, \emph{what properties of the representation space determine which data are discovered earlier?}

Prior work has evaluated embeddings through separability, local purity, and clustering quality \citep{rousseeuw1987silhouettes,davies1979cluster,schnabel2015evaluation}. Related work on \emph{instance hardness} links local class overlap to classification difficulty, including through neighborhood-based measures such as $k$-Disagreeing Neighbors \citep{smith2014instance}. However, these works primarily study difficulty for \emph{classification}. In fine-grained ecological workflows where the goal is to transductively label a fixed pool, we instead ask which points are geometrically difficult to \emph{sample} early during active labeling.

To study this phenomenon, we introduce a notion of \emph{sampling difficulty}, which measures how likely a sample is to remain hidden during active labeling. Intuitively, samples are harder to discover when they lie in dense regions of the embedding space but are surrounded by examples from other classes.

For each sample $i$, we define its local neighborhood using clusters induced by $k$-means in the embedding space. Specifically, we run $k$-means over the full deployment pool $\mathcal{D}$ using the embeddings $z=f_\theta(x)$, and let $c(i)$ denote the cluster assignment of sample $i$. We then define the neighborhood of $i$ as $\mathcal{C}(i)=\{j \in \mathcal{D}: c(j)=c(i),\, j\neq i\}.$ 
Here, $k$ denotes the number of $k$-means clusters and determines the granularity at which local structure is measured. Unless otherwise noted, we set $k$ to the number of ground-truth classes in the dataset. In Appendix~\ref{app:difficulty_robustness}, we ablate over clustering algorithm and cluster counts around the number of true classes and show that the identified needle sets are relatively stable across these choices. We use these cluster-based neighborhoods to characterize two aspects of discoverability: \textit{local density}---how densely a sample is embedded---and \textit{local isolation}---how isolated it is with respect to its own class.

\paragraph{(1) Local density.}
We measure how dense a sample's neighborhood is by first computing the average pairwise distance across the full deployment pool,
\begin{equation}
\mu_{\mathrm{pool}}=
\frac{2}{|\mathcal{D}|(|\mathcal{D}|-1)}
\sum_{i<j}\|z_i-z_j\|_2,
\end{equation}
which provides a reference distance scale for the embedding space. For each sample $i$, we then compute the average distance from $i$ to other samples in its assigned cluster,
\begin{equation}
\mu_{\mathrm{clust}}(i)=
\frac{1}{|\mathcal{C}(i)|}
\sum_{j\in \mathcal{C}(i)}\|z_i-z_j\|_2.
\end{equation}
Local density is defined by the ratio 
\begin{equation}
    \rho_i = \frac{\mu_{\mathrm{pool}}}{\mu_{\mathrm{clust}}(i)}.    
\end{equation}
Large values of $\rho_i$ indicate that the sample lies in a locally dense region of the representation space. This is related in spirit to density-aware active learning  \citep{zhu2008active,settles2009active}, but here density is used to analyze discoverability rather than to define a query rule. In Appendix \ref{app:difficulty_robustness}, we verify that the identified needle sets are stable across distance metric choice.

\paragraph{(2) Local isolation.}
We measure how well a sample is locally surrounded by others from the same class within its assigned cluster:
\begin{equation}
\pi_i=
\frac{1}{|\mathcal{C}(i)|}
\sum_{j\in\mathcal{C}(i)} \mathbf{1}[y_j=y_i].
\end{equation}
This quantity is the fraction of samples in the same cluster that share the label of sample $i$. High values of $\pi_i$ indicate that the sample is well isolated with respect to its own class, while low values indicate that it belongs to a cluster dominated by other classes. This is closely related to overlap-based hardness measures such as $k$-Disagreeing Neighbors \citep{smith2014instance}, but uses cluster-induced neighborhoods rather than fixed-size nearest-neighbor sets.

We combine these two quantities to define the sampling difficulty,
\begin{equation}
    d_i=\frac{\rho_i}{\pi_i}.
\end{equation}
Difficulty is therefore high when a point lies in a dense region but has low local isolation. Such points are geometrically hidden, like \textit{needles in a haystack}. Unlike prior density or overlap measures, which are typically used to explain classification difficulty or guide representative sampling, our score is intended to quantify \emph{discoverability}: whether a sample is likely to be surfaced early during fixed-pool active labeling.

Since $d_i$ is continuous and its scale may vary across datasets, we define \emph{needles} as the upper tail of the difficulty distribution within each dataset rather than through a fixed absolute threshold; unless otherwise noted, we use the top 25\% most difficult samples for the analysis and visualizations below. Our conclusions do not depend on the exact percentile used; results are qualitatively stable across a range of high-percentile thresholds (see Appendix \ref{app:difficulty_robustness}).

\begin{figure}[h]
    \vspace{-4mm}
    \centering
    \includegraphics[width=\linewidth]{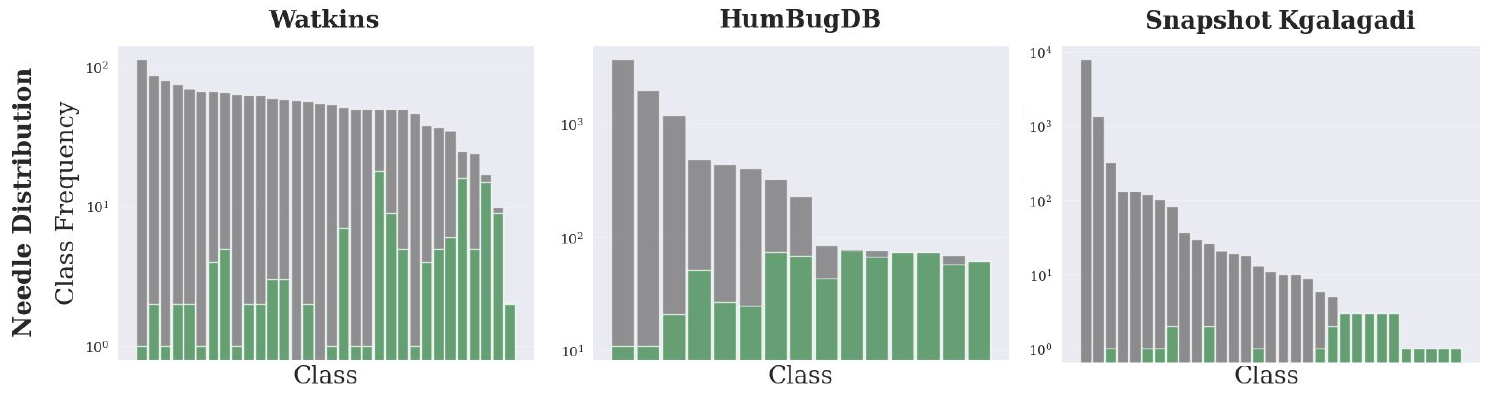}
    \caption{\textbf{Needles typically concentrate in the long tail, across datasets.} Across datasets, needles are both skewed toward minority classes and systematically harder to sample due to dense, mixed local neighborhoods in embedding space.}
    \label{fig:needles_difficulty}
    \vspace{-4mm}
\end{figure}

\begin{figure}
    \centering
    \includegraphics[width=0.8\linewidth]{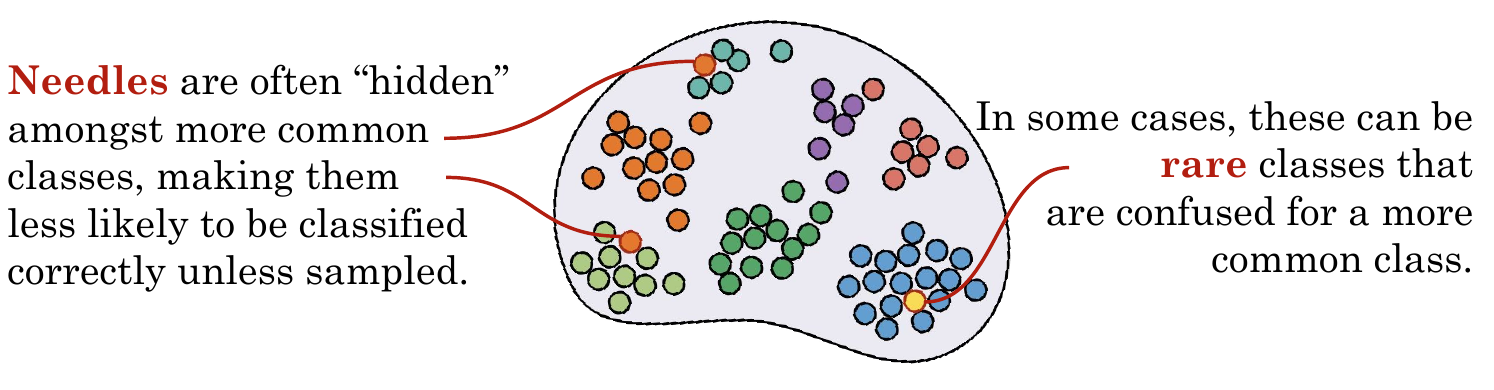}
    \caption{\textbf{Needles are minority-class samples embedded within or adjacent to dense clusters of dominant classes.} Because acquisition strategies operate on embedding geometry rather than labels, such points are less likely to be sampled early despite being scientifically important, in some cases.}
    \label{fig:needles_explanatory}
    \vspace{-5mm}
\end{figure}

Acquisition strategies operating on density or uncertainty are likely, simply due to local class imbalance, to select the more abundant class in a region, delaying discovery of these high-difficulty points. We show in Figure~\ref{fig:needles_difficulty} that needles are concentrated in the long tail (analyses on additional datasets are shown in Appendix Figure \ref{fig:needle_dist_ext}). However, needles are not \textit{always} rare categories. Under our definition, a class may be globally rare yet geometrically isolated and therefore easy to discover via standard uncertainty sampling. Conversely, a sample may be difficult to sample because it lies within a dense region dominated by another class, even if its class is not extremely rare overall. Figure~\ref{fig:needles_explanatory} illustrates this concept.

To contextualize this in the bioacoustic domain, a single species may produce acoustically distinct vocalizations depending on age, behavioral context, recording device, or environmental conditions. Adult and juvenile calls, for example, often form separate subclusters in embedding space due to differences in pitch and temporal structure. Similarly, alarm calls, mating calls, and contact calls can occupy distinct acoustic regions despite sharing the same taxonomic label.

When dominant species exhibit such intra-class variability, some subclusters may overlap with other semantically similar classes. Rare taxa can therefore become embedded within dense regions associated with common species.

\paragraph{Difficult Data Are Discovered Late.}
To understand the dynamics of rare class discovery in transductive active labeling, we analyze the relationship between {discovery time}---the active learning cycle at which a data point is sampled--- and {needle difficulty}.

\begin{figure}[h]
    \centering
    \includegraphics[width=\linewidth]{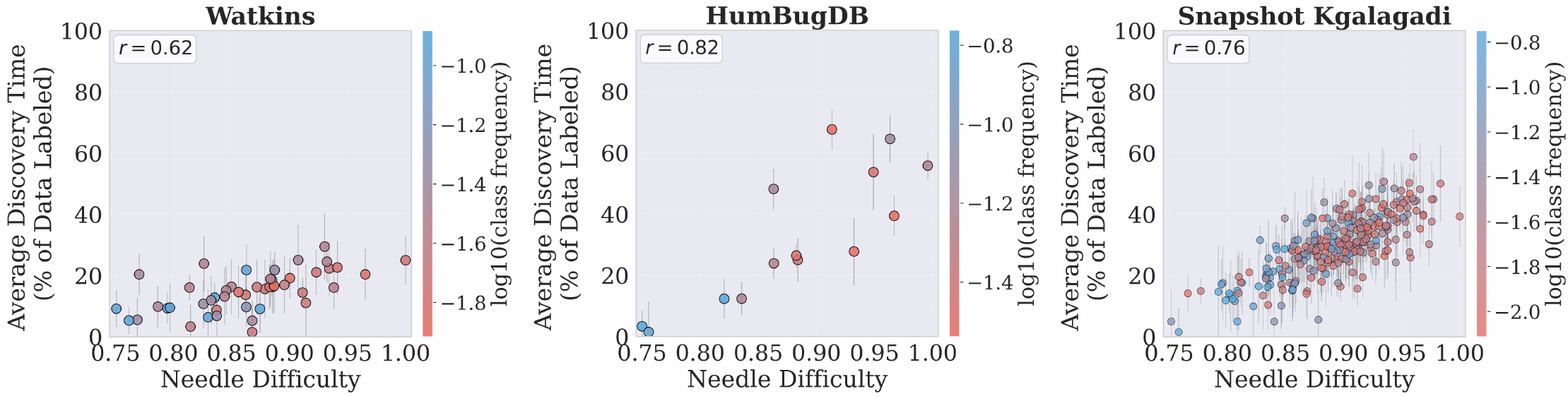}
    \caption{\textbf{Difficult data are discovered later, across datasets.} We compare the difficulty of each needle against the cycle at which it is sampled, averaged across sampling strategies.}
    \label{fig:difficulty_over_time}
\end{figure}

Figure~\ref{fig:difficulty_over_time} confirms a clear positive relationship between sample difficulty and discovery time on multiple datasets: data embedded within dense, mixed neighborhoods are systematically discovered later during active learning. This pattern holds across datasets (see Appendix Figure \ref{fig:discovery_other}). Although data from rarer classes tend to exhibit higher difficulty, the relationship is not strictly uniform. Some rare data are geometrically isolated and discovered early, while others are embedded within dominant class data and remain hidden until much later.

\paragraph{The Difficulty Landscape Is Dynamic.}
\label{sec:difficulty_dynamics}

\begin{figure}[h]
    \centering
    \includegraphics[width=0.95\linewidth]{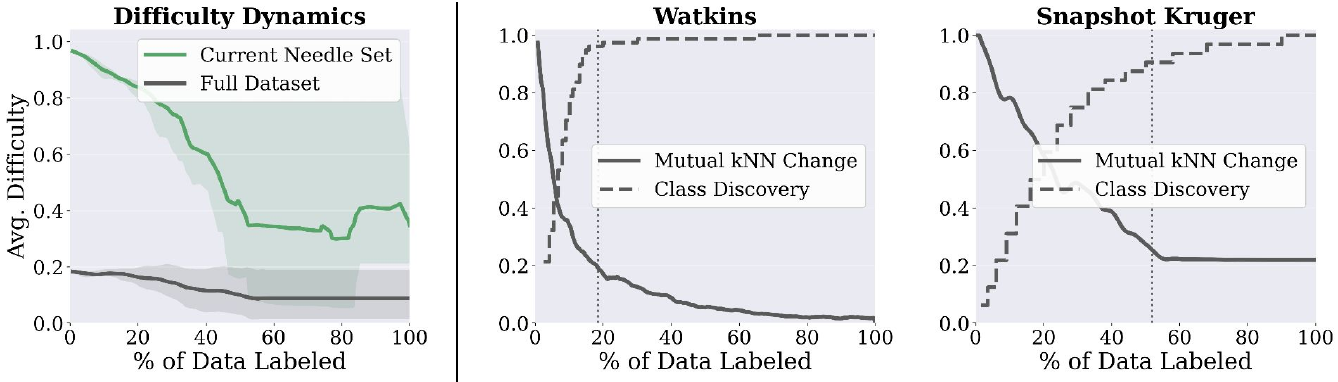}
    \caption{\textbf{Sampling difficulty evolves as active labeling reshapes the geometry of the pool.} Left: the average difficulty of both needles and the full pool decreases as labeling progresses, across datasets. Middle: on Watkins, representation change (as measured by a mutual kNN between cycles) stabilizes as class discovery saturates. Right: on the more long-tailed Snapshot Kruger dataset, the representation stabilizes later, and class coverage is slower.}
    \label{fig:difficulty_dynamics}
\end{figure}

Sampling difficulty changes during active labeling. Figure~\ref{fig:difficulty_dynamics} shows that active labeling makes the discovery problem easier over time. Average difficulty over the full dataset and needle set at each cycle both decline, indicating that labeling removes hard examples \textit{and} changes the geometry of the remaining pool.

On the Watkins Marine Mammal dataset, the representation shifts most early in labeling, when class discovery also rises fastest. As discovery saturates, the representation stabilizes. Thus, many needles are not intrinsically hard. They are hard under the current geometry, and become easier once nearby informative examples are labeled and the representation is updated. Snapshot Kruger shows the same pattern in a more discovery-limited, long-tailed setting: stabilization occurs later, when class coverage begins converging.

Needle difficulty is therefore representation-dependent, not fixed. Representation quality is shaped by several factors, including the label ontology and the embedding model itself.  

Coarsening the label set by merging similar categories can make the embedding space appear better separated, because the task no longer requires resolving fine-grained distinctions. Conversely, finer-grained ontologies expose boundaries that may be collapsed or poorly separated in the representation. On Watkins, grouping species into broader marine-mammal categories reduces the number of needles from 47 to 6, an 87.2\% reduction, while increasing grouped class-averaged accuracy from 0.762 to 0.956. ReefSet shows the same trend: grouping reef sounds into broad acoustic-source categories reduces needles from 1,357 to 265, an 80.5\% reduction, and increases grouped class-averaged accuracy from 0.936 to 0.994. On Snapshot Kruger, collapsing species to an order-level ontology reduces needles from 248 to 73, a 70.6\% reduction, and increases grouped class-averaged accuracy from 0.803 to 0.940. These results show that long-tail difficulty is not only a function of class frequency; it is also a function of task granularity, and a coarse label ontology can reduce the number of geometrically difficult examples.

The embedding model, on the other hand, directly determines the geometry of the space over which active labeling operates. Across bioacoustic datasets, stronger encoders produce higher cluster alignment, fewer needles, and earlier recovery of rare samples under the same annotation budget. Perch 2.0 consistently yields the strongest cluster alignment and the fewest needles among the evaluated audio encoders. Averaged across datasets, Perch 2.0 improves NMI from 0.51 to 0.59 relative to Perch 1.0, reduces the average needle count by about 8\%, and increases rare-sample recovery from 0.58 to 0.65. The same pattern holds more strongly relative to weaker or less specialized encoders, which leave more rare examples geometrically obscured. These results indicate that improving representation geometry can substantially reduce discovery difficulty, sometimes as much as changing the acquisition heuristic itself. We examine the effect of embedding model choice thoroughly in Appendix \ref{app:embedding_geometry}. 

\vspace{5pt}\noindent \textbf{When to Stop Labeling.}
In transductive active labeling, strong predictive performance on common classes is necessary but not sufficient; active labeling must also continue long enough to discover rare classes. In long-tailed datasets, this makes \emph{discovery} a central stopping signal. 

This perspective is closely related to \emph{ecological rarefaction} \citep{raup1975taxonomic}, which plots the cumulative number of unique species observed as a function of field sampling effort. Sampling is considered effectively complete when the rarefaction curve plateaus and additional sampling yields few or no newly discovered species. Transductive active labeling has an analogous structure: as experts label more examples, the system should eventually stop discovering new classes. However, unlike classical rarefaction, active labeling must balance two goals simultaneously: discovering the rare tail and achieving accurate predictions on the remaining unlabeled pool.

In practice, experts decide when to stop labeling using a fixed labeling budget \citep{settles2009active}, pseudo-label consistency across active-learning cycles \citep{bloodgood2009method,ghayoomi2010using}, or predictive performance on a held-out labeled validation set \citep{pullar2024hitting,dumoulin2025search}. 

\textbf{Prediction Stability} measures by pseudo-label changes on the remaining unlabeled pool
\begin{equation}
C_t = \frac{1}{|U_t|} \sum_{x \in U_t} \mathbf{1}\!\left[\hat y_t(x) \neq \hat y_{t-1}(x)\right].
\end{equation}
Low values of $C_t$ indicate that the model is no longer substantially changing its predictions over the unlabeled pool \citep{bloodgood2009method,ghayoomi2010using}.

An \textbf{Inductive Threshold} stops labeling once predictive performance on a held-out labeled set exceeds a target value. Formally, if $A_t$ denotes a held-out performance metric at active-learning cycle $t$, then inductive stopping declares convergence when $A_t \geq \tau_{\mathrm{ind}},$ where $\tau_{\mathrm{ind}}$ is a predefined performance target. 

To capture the importance of discovery in transductive labeling, we introduce a rarefaction-inspired \textbf{Discovery Stability} criterion, which asks whether successive annotation batches are still revealing new classes. Let $d_t$ denote the fraction of newly observed classes in batch $t$. We smooth this signal over the most recent $w$ active-learning cycles $\bar d_t = \frac{1}{w}\sum_{i=t-w+1}^{t} d_i.$ Discovery is considered saturated when
$\bar d_t < \tau_{\text{disc}}.$ In relatively simple datasets, class discovery may plateau early even though additional labels still improve classifier quality; rare discoveries can also be intermittent, so a brief period without new classes being sampled does not necessarily imply that the remaining pool has been exhausted.

We propose a \textbf{Hybrid Stopping Rule} that stops only after \textit{both} predictive performance and discovery saturation conditions have been satisfied. Concretely, this rule stops at $\max(t_{\mathrm{ind}}, t_{\mathrm{disc}})$. This criterion ensures the two goals of transductive active labeling are met: achieving strong predictive performance on the dominant structure while continuing long enough to recover rare, scientifically important classes.

\begin{figure}[t]
    \centering
    \includegraphics[width=0.85\linewidth]{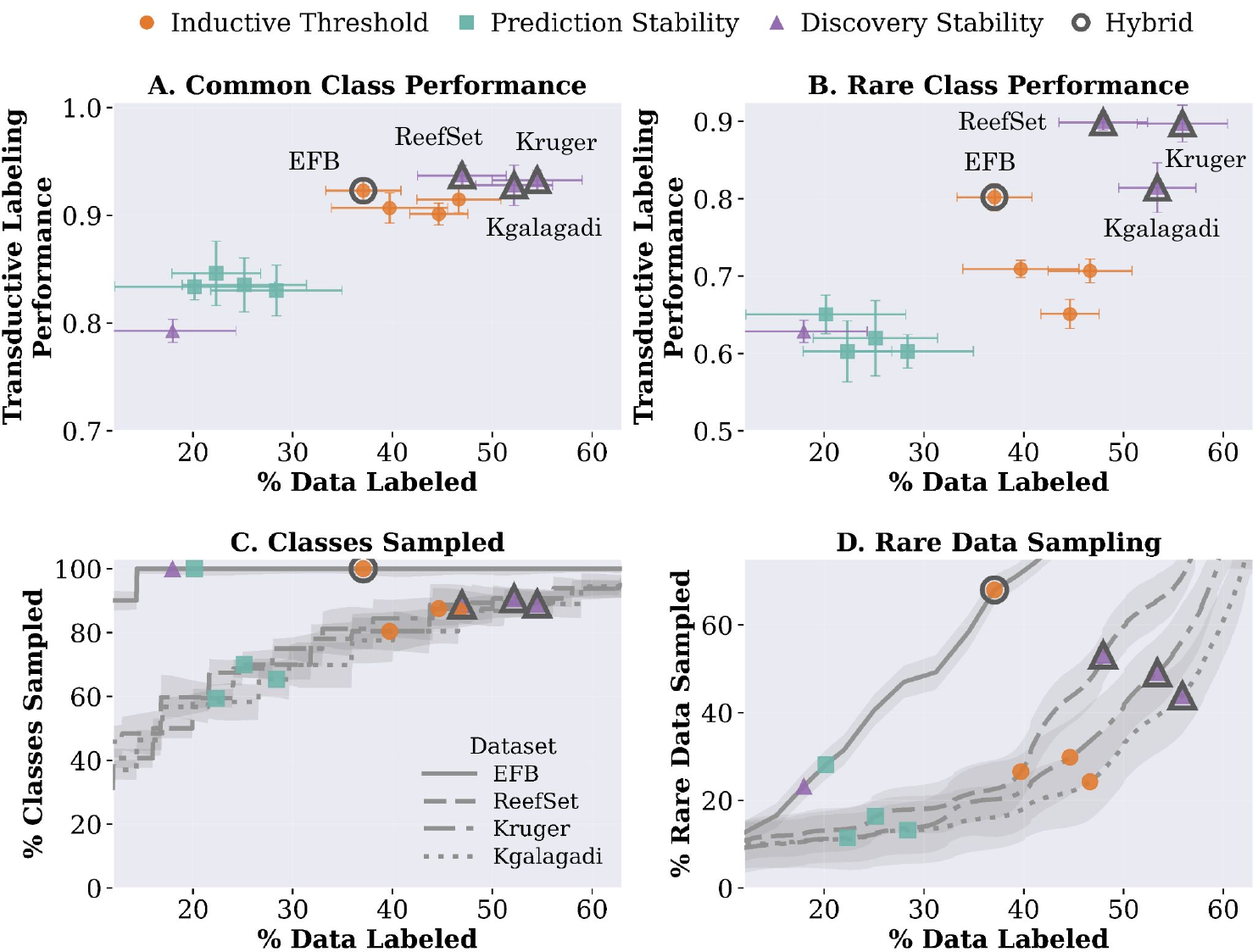}
    \caption{\textbf{Our proposed hybrid stopping rule balances predictive performance, discovery, and label efficiency.} Each point shows the fraction of the pool labeled and transductive performance at stopping, with error bars across strategies. On long-tailed datasets, discovery stability labels more data, but improves rare class recovery.}
    \label{fig:stopping}
\end{figure}

Figure~\ref{fig:stopping} highlights a consistent tradeoff between label efficiency and performance on the rare tail. We compare the four stopping rules described above: {prediction stability}, {inductive threshold}, {discovery stability}, and the {hybrid} criterion. We use a randomized dataset-wise tuning split (CBI, Dogs, Watkins, HumBugDB, Camdeboo) to select stable parameters (see Appendix~\ref{app:sensitivity}), and report final stopping performance on held-out datasets (EFB, ReefSet, Kruger, Kgalagadi). In our experiments, we choose conservative values that lie in stable regions of the sensitivity analysis. We set $\delta=0.01$ as a conservative prediction-stability threshold: smaller values delay stopping until pseudo-label entropy is very low, but yield only modest gains. We set $w=6$ to smooth the discovery signal; smaller windows are highly sensitive to 
brief periods without new class sampled, while larger windows add little stability. We set $\tau_{\mathrm{disc}}=0.005$ because it lies in a broad stable region. We set $\tau_{\mathrm{ind}}=0.95$, following prior agile modeling work \cite{dumoulin2025search}, because higher ROC-AUC thresholds mainly improve performance by requiring substantially more labels. Finally, we use a 20\% held-out set because smaller validation sets make inductive stopping noisy, while larger sets add limited stability and remove more labeled examples from training. Importantly, for inductive stopping, we adjust the reported fraction labeled to include the validation set, since these examples also require expert annotation. These stopping parameters are tunable and should be adapted to the task. For example, when the test distribution is known to be representative and predictive performance is the primary objective, the inductive threshold can be increased. 

Prediction stability consistently stops early, and is therefore the most label-efficient rule, but this efficiency comes at a clear cost in performance. In contrast, inductive thresholding is strongest on simpler, balanced datasets such as EFB, where all classes are sampled early on and additional labels mainly improve the classifier. Inductive accuracy remains a reliable proxy for predictive performance over the unlabeled pool. In our transductive setting, the labeled examples used to construct the validation set for inductive stopping are not wasted effort: they still contribute to the overall goal of labeling the fixed pool.

On more long-tailed, discovery-limited datasets like ReefSet, Kruger, and Kgalagadi, discovery stability yields substantially better rare-class performance, with a slightly larger labeling budget, sampling more of the data in the long tail than inductive threshold-based stopping. Inductive evaluation is a strong proxy for predictive performance, but it is not always the right target in a transductive setting. A model can achieve high predictive performance on the dominant structure in the pool while rare, but potentially scientifically important categories remain unsampled. The hybrid rule provides the most practical default: it remains efficient on simpler pools while avoiding premature stopping on long-tailed datasets where discovery is the dominant bottleneck.

\vspace{-5pt}
\section{Conclusion}
Many ecological annotation workflows are inherently \emph{transductive}, where the goal is to accurately label a collected pool of data. Yet most methods are evaluated \textit{inductively}, ignoring the human-in-the-loop and misrepresenting real annotation performance. We empirically demonstrate across diverse bioacoustic and image datasets that this gap is most exaggerated in the long tail of the data distribution. Because classifiers struggle with rare classes but experts label them easily once surfaced, performance becomes \textit{discovery-limited:} the key challenge is sampling those examples to label.

This is easy when the embeddings of rare categories are well-separated in the representational geometry of the dataset, but far more difficult when they are isolated within dense dominant-class regions. Our analysis formalizes this as the notion of ``finding needles in haystacks,'' and we show that such samples are discovered systematically later during labeling. Future work should benchmark methods as \emph{transductive active labeling} problems rather than inductive active learning tasks, and continue building on discovery-centered sampling strategies designed to surface rare and informative examples. 

In practice, we cannot measure transductive performance on unlabeled data, making it unclear when to stop labeling. To address this, we introduce a stopping criterion inspired by ecological rarefaction that tracks the declining yield of samples. This criterion consistently improves performance on rare classes while maintaining strong labeling performance. Our proposed shift of focus towards transductive labeling better matches the needs of ecologists, and will hopefully lead to the development of methods better-targeted to the reliable and efficient annotation of large-scale environmental datasets.

\bibliographystyle{plainnat} 
\bibliography{main} 

\newpage
\input{appendix}

\end{document}

%% file: appendix.tex
\section{Supplementary Material}

\subsection{Audio Preprocessing}

All bioacoustic datasets undergo standardized preprocessing to ensure consistent input for embedding extraction. Raw audio files are converted into fixed-length 5-second clips using the following protocol: (1) clips shorter than 5 seconds are zero-padded at the end to reach the target duration, (2) clips longer than 5 seconds are trimmed by extracting the highest-energy 5-second window using a sliding window energy calculation, and (3) clips of exactly 5 seconds are retained as-is. This preprocessing ensures uniform temporal dimensions while preserving the most informative acoustic content from longer recordings. The processed clips are stored with their corresponding class labels in CSV format for downstream embedding extraction and active learning experiments. 

\subsection{Classifier Implementation Details}
\label{app:classifier_details}

For all main experiments, we train a lightweight classifier on top of frozen embeddings. After each acquisition cycle, the classifier is retrained on the full set of labeled examples accumulated so far.

Our default classifier is a multinomial logistic-regression probe implemented with \texttt{LogisticRegression} from scikit-learn. Given a $d$-dimensional input embedding $z \in \mathbb{R}^{d}$, the classifier predicts class logits
\[
\ell = zW + b,
\]
where $W \in \mathbb{R}^{d \times C}$ and $b \in \mathbb{R}^{C}$ for a dataset with $C$ classes. This corresponds to a single linear layer with no hidden units. The model is trained with multinomial cross-entropy loss. We use scikit-learn's default \texttt{lbfgs} solver with L2 regularization, \texttt{C=1.0}, \texttt{max\_iter=500}, and a fixed \texttt{random\_state} for reproducibility.

At active learning cycle $t$, the classifier is fit on all labeled examples in $L_t$ and then used to produce class probabilities over the unlabeled pool. These probabilities are used by uncertainty-based acquisition strategies such as least-confidence and margin sampling, and the resulting predictions are used for both inductive and transductive evaluation. We retrain the classifier from scratch after each cycle rather than warm-starting from the previous round.

\paragraph{MLP comparison.}
We also evaluate an MLP baseline to test whether our conclusions depend on using a strictly linear classifier. This model is implemented with scikit-learn's \texttt{MLPClassifier} and uses a single hidden layer with 64 units. It is trained with cross-entropy loss, \texttt{max\_iter=500}, default L2 regularization (\texttt{alpha=0.0001}), and no early stopping. Unless otherwise noted, all other aspects of the active learning pipeline remain unchanged.

\begin{figure}[h]
    \centering
    \includegraphics[width=\linewidth]{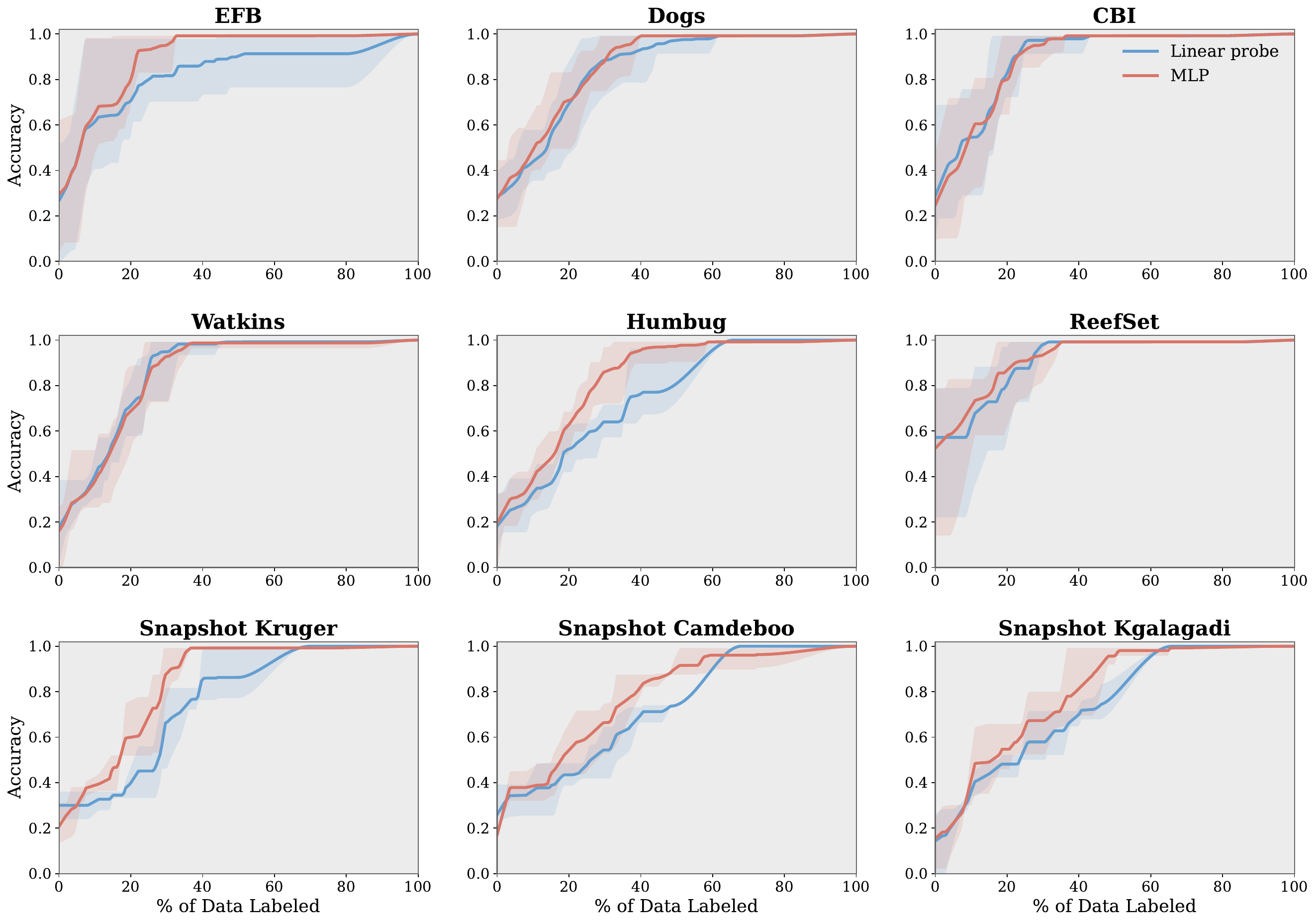}
    \caption{\textbf{Linear probe vs. MLP performance across datasets.} For each dataset, we compare accuracy as a function of labeling budget for a linear probe and a multilayer perceptron (MLP) trained on the same learned representation. Solid lines show the mean trajectory across active learning strategies, and shaded bands indicate the strategy-wise range. Across datasets, the MLP generally outperforms the linear probe, with especially large gains on the image datasets (Snapshot Kruger, Camdeboo, and Kgalagadi), suggesting that representation quality can determine classifier expressivity, especially when the initial embeddings are not linearly separable.}
    \label{fig:watkins_strat}
\end{figure}

The goal of this work is to study the effect of sample selection and evaluation protocol in realistic fixed-pool labeling workflows, rather than to optimize classifier capacity on top of frozen embeddings. Our use of simple linear classifiers is consistent with standard practice in prior bioacoustic embedding-based pipelines, where lightweight probes are often used. Moreover, for the foundation-model embeddings considered here, a simple classifier is already sufficient to separate many classes reasonably well, allowing us to focus on the role of sample selection rather than classifier expressivity.

\subsection{Robustness of Sampling Difficulty Definition}
\label{app:difficulty_robustness}

Our main analysis defines sampling difficulty using a geometric, label-relative criterion. Intuitively, a difficult sample is one whose embedding lies in a locally ambiguous region with few nearby samples from the same class. This definition is intended to capture samples that are hard to discover through standard training dynamics: they are not merely rare in the dataset, but rare in the representation geometry in a way that limits local class evidence. Because this score depends on choices such as the neighborhood construction, distance metric, and difficulty threshold, we evaluate whether our conclusions are robust to these design decisions.

\begin{figure}[h]
    \centering
    \includegraphics[width=\linewidth]{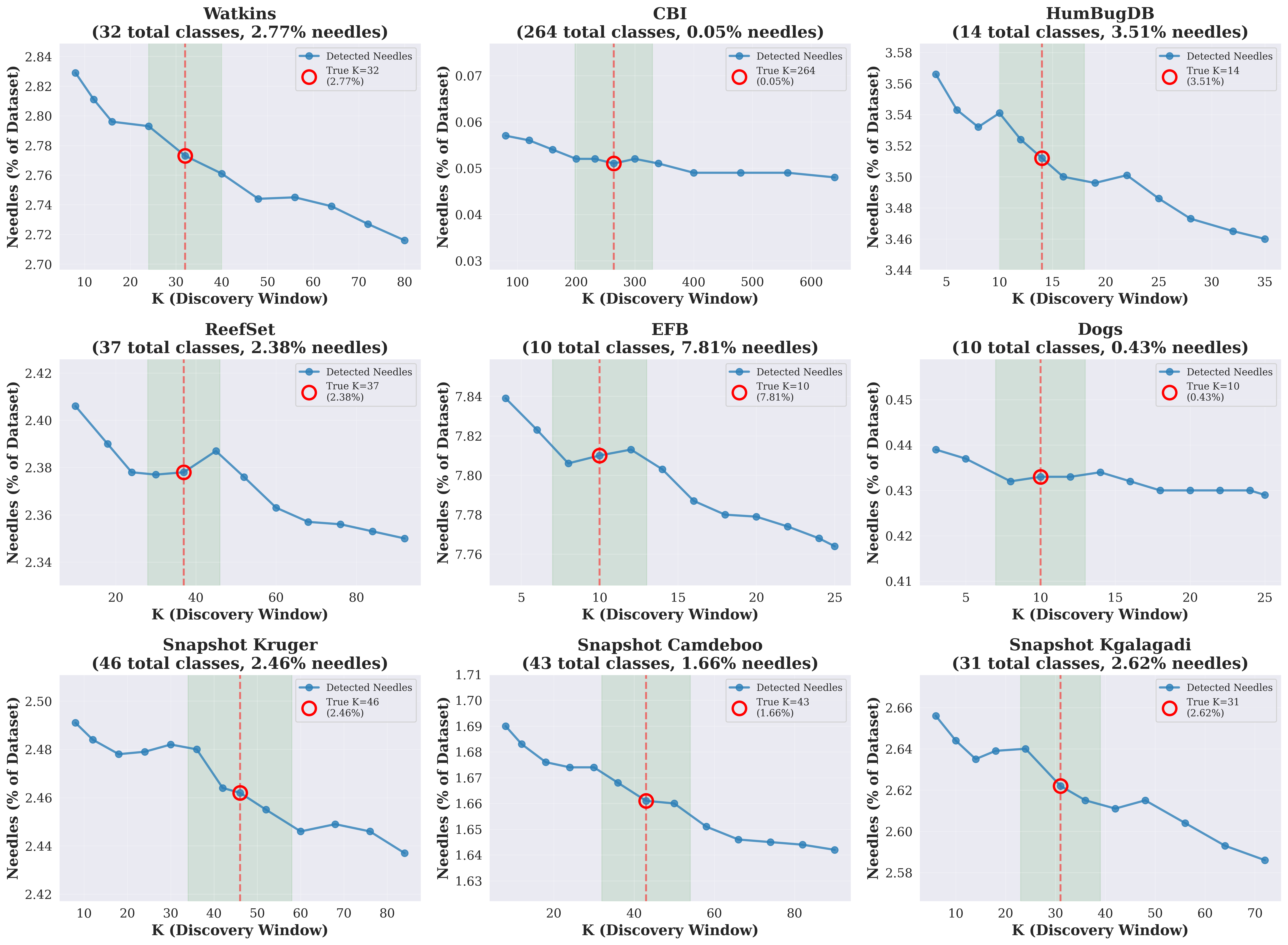}
    \caption{\textbf{Sensitivity of needle detection to the number of clusters $K$ used to compute sampling difficulty.} We evaluate how the number of clusters used in $K$-means affects the identification of rare-class samples (“needles”). For each dataset, we vary $K$ and measure the number of needles detected under our geometric difficulty criterion. The dashed red line indicates the value of $K$ equal to the true number of classes in the dataset, while the green band highlights the neighborhood around this value. As expected, the number of detected needles depends on $K$, since clustering resolution determines how finely the embedding space is partitioned. However, the curves change smoothly and remain stable near the true class count. We therefore use the true number of classes as the clustering resolution in our analysis, as it provides a natural and interpretable scale for approximating semantic structure in the embedding space while avoiding arbitrary hyperparameter choices.}
    \label{fig:k_sensitivity}
\end{figure}

\paragraph{Sensitivity to clustering resolution.}
We use $k$-means to define local neighborhoods for measuring density and label isolation, setting $k$ to the true number of classes in each dataset. This choice provides a natural semantic scale: it partitions the embedding space at approximately the granularity of class-level structure, without introducing an additional tuned hyperparameter. Appendix Fig.~\ref{fig:k_sensitivity} evaluates sensitivity to this choice by varying the number of clusters $K$ and measuring the resulting number of detected needles. As expected, needle counts depend on $K$, since finer partitions expose smaller low-density regions and coarser partitions merge distinct semantic neighborhoods. However, the curves vary smoothly and remain stable in a neighborhood around the true class cardinality. This suggests that our results are not driven by a brittle choice of clustering resolution, but by a persistent geometric signal near the relevant semantic scale.

\paragraph{Robustness to neighborhood construction and distance metric.}
Table~\ref{tbl:robustness_table} compares our default definition against several perturbations. We report three quantities: rank correlation $\rho$ between difficulty scores, Jaccard overlap between the top-25\% most difficult samples and the default needles, and the correlation $r$ between difficulty and sampling/discovery time. These metrics separate two notions of robustness. Rank correlation asks whether the overall ordering of samples is preserved; Jaccard overlap asks whether the same high-difficulty subset is identified; and sampling-time correlation asks whether the resulting notion of difficulty preserves the behavioral phenomenon of interest.

Agglomerative clustering behaves similarly to $k$-means, producing high rank correlations, substantial overlap among top-difficulty samples, and comparable correlations with delayed discovery. This indicates that the signal is not specific to the optimization of $k$-means. Replacing Euclidean distance with cosine distance also yields similar needle sets, suggesting that the embedding geometry is stable under standard choices of metric. In contrast, HDBSCAN produces only moderate score agreement and substantially lower overlap with the default needles. More importantly, the HDBSCAN-derived needles are weakly correlated with later discovery, suggesting that its density-adaptive clusters capture a different notion of outlierness rather than the sampling difficulty signal we study. We therefore use the simpler class-scale clustering construction, which better aligns with delayed discovery.

\begin{table}[h]
\centering
\scriptsize
\setlength{\tabcolsep}{4pt}
\renewcommand{\arraystretch}{0.86}
\begin{tabular}{llccc}
\toprule
\textbf{Category} & \textbf{Variation} & \textbf{$\rho$} & {Top-25\% Jac.} & \textbf{$r$} \\
\midrule
\multirow{2}{*}{Clustering}
  & Agglomerative  & .76--.91 & .60--.74 & .52--.76 \\
  & HDBSCAN & .45--.63 & .28--.46 & .08--.21 \\
\midrule
Distance
  & Cosine vs. Euclid. & .87--.95 & .70--.77 & .46--.69 \\
\midrule
\multirow{2}{*}{Alt. difficulty}
  & LID & .54--.68 & .34--.49 & .24--.41 \\
  & Intrinsic Curvature & .18--.32 & .12--.24 & .03--.17 \\
\bottomrule
\end{tabular}
\caption{\textbf{Difficulty robustness across datasets.} We compare the default sampling-difficulty score to perturbations of the neighborhood construction, distance metric, and geometric difficulty definition. We report score rank correlation $\rho$, Jaccard overlap with the default top-25\% difficult samples, and correlation $r$ between difficulty and discovery time.}
\label{tbl:robustness_table}
\end{table}

\begin{figure}[h]
    \centering
    \includegraphics[width=\linewidth]{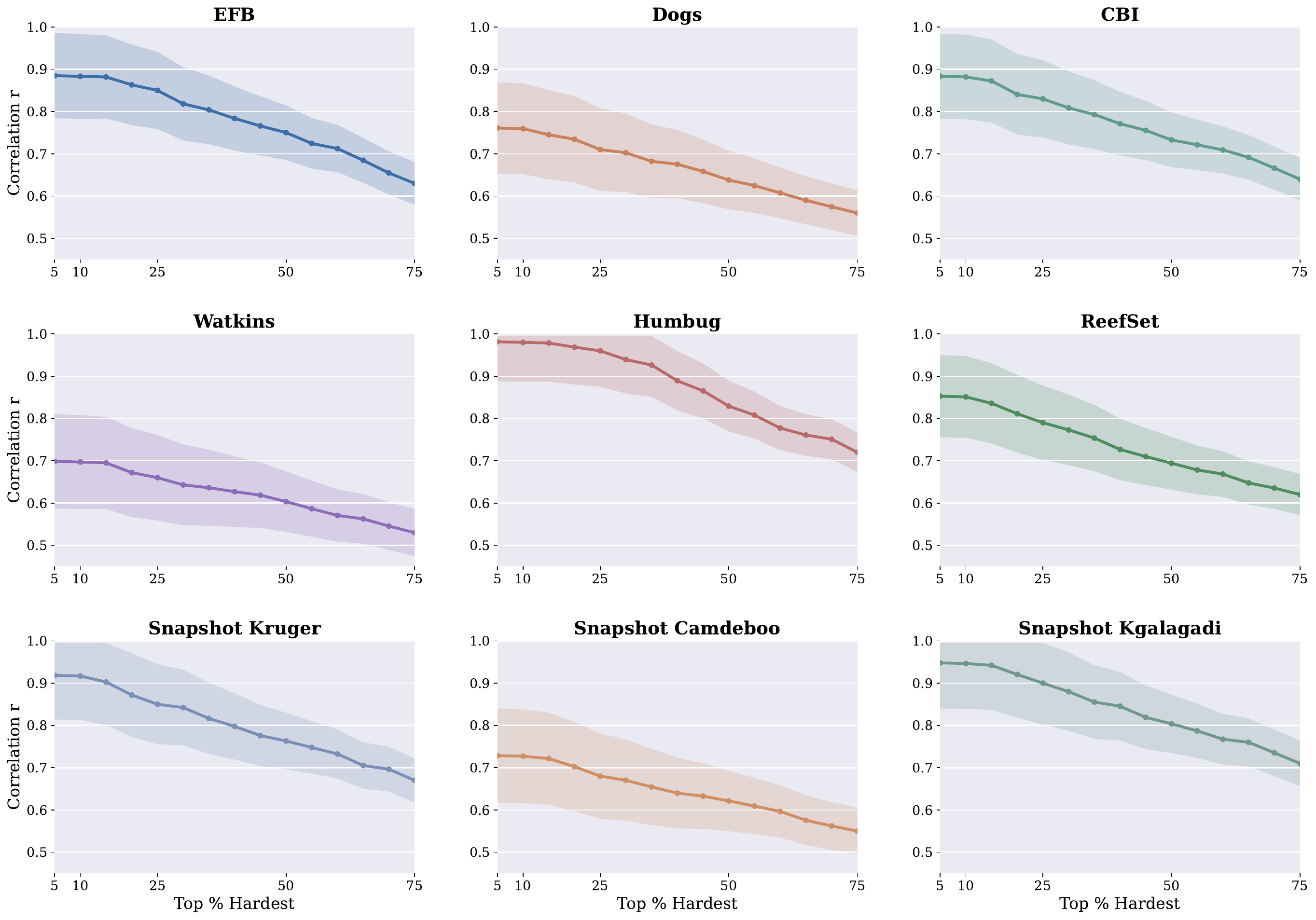}
    \caption{\textbf{Harder needles are discovered later, across difficulty thresholds.} For each dataset, we vary the threshold used to define the ``hard'' subset from the top 5\% to the top 75\% of samples ranked by difficulty and measure the correlation $r$ between difficulty score and discovery time under active sampling. Across datasets, the relationship remains positive over a broad range of thresholds, indicating that harder samples are consistently discovered later. The effect is strongest for the most extreme difficult samples, especially around the top 5--10\%, and weakens gradually as the threshold is relaxed to include more borderline cases, but it does not disappear.}
    \label{fig:threshold_ablation}
\end{figure}

\paragraph{Robustness to the difficulty threshold.}
Our main experiments define needles using the top 25\% of the difficulty distribution. This threshold is not meant to imply a sharp phase transition between difficult and easy samples; rather, it provides a fixed operating point for comparing datasets. To verify that the qualitative conclusions are not threshold-specific, we repeat the analysis for thresholds ranging from the top 5\% to the top 75\% of samples. Across this range, the same trends persist: high-difficulty samples are disproportionately concentrated in the long tail, and they are discovered later during sampling. The relationship between difficulty and discovery time remains positive across thresholds, with correlations in the range $r=.48$--$.71$. Figure \ref{fig:threshold_ablation} shows that the effect is strongest for the most extreme samples, particularly in the top 5--10\%, consistent with the interpretation that the highest-difficulty region contains the clearest ``needle'' examples. As the threshold is relaxed, the difficult set includes more borderline cases, so the magnitude of the effect decreases but does not disappear.

\paragraph{Comparison to alternative geometric difficulty scores.}
We also compare against two alternative geometry-only diagnostics: local intrinsic dimensionality (LID) and intrinsic curvature. LID partially recovers the delayed-discovery signal, yielding moderate rank correlations and nontrivial overlap with our default difficult set. This suggests that local expansion or sparsity in the embedding space is indeed one component of sampling difficulty. However, LID does not explicitly account for class support, so it cannot distinguish a geometrically sparse but class-consistent region from a sparse region that is label-isolated. Intrinsic curvature is substantially weaker across all metrics, with low overlap and weak correlation with sampling time. This indicates that curvature alone does not capture the class-relative scarcity that determines whether a sample is difficult to discover.

These ablations support two conclusions. First, the empirical findings in our paper are robust to reasonable perturbations of clustering algorithm, distance metric, and difficulty threshold. Second, the strongest predictor of delayed discovery is not generic geometric atypicality alone, but a label-relative notion of geometric difficulty that combines local density with class isolation. This motivates our use of the proposed sampling-difficulty score in the main analysis and suggests that related label-aware geometric diagnostics may be useful for future active-sampling methods.

\subsection{Exploring the Effect of Embedding Model Choice}
\label{app:embedding_geometry}

Changing the label ontology affects \emph{which} distinctions must be resolved; changing the embedding model affects \emph{how separable} those distinctions are in the first place. Together, these factors determine the geometry over which transductive active labeling operates. If rare classes are geometrically isolated, most acquisition rules can surface them efficiently; if they remain entangled with dominant structure, even strong heuristics struggle to recover them quickly. In this sense, embedding quality directly influences how much data must be labeled before the long tail becomes accessible.

To quantify this effect, we compare embedding models across datasets under a fixed annotation budget using margin sampling. For each dataset-model pair, we report three complementary quantities: cluster alignment (NMI), the number of needles, and the fraction of rare samples surfaced during active labeling. Tables~\ref{tab:appendix_embedding_nmi}--\ref{tab:appendix_embedding_rare} show a consistent pattern: stronger encoders produce better cluster alignment, fewer needles, and earlier recovery of rare examples. The effect is especially pronounced on discovery-limited datasets such as ReefSet and the camera-trap datasets, where representation quality has a large influence on how quickly rare structure becomes separable.

\begin{table*}[t]
\centering
\caption{\textbf{Cluster alignment (NMI) across datasets and embedding models.} Higher is better. SurfPerch is excluded from ReefSet comparison because of overlap with the training dataset of this model.}
\label{tab:appendix_embedding_nmi}
\tiny
\setlength{\tabcolsep}{5pt}
\begin{tabular}{lcccccc}
\toprule
\textbf{Dataset} & \textbf{Perch 2.0} & \textbf{Perch 1.0} & \textbf{SurfPerch} & \textbf{BirdNET} & \textbf{BEATS} & \textbf{AudioMAE} \\
\midrule
Watkins     & \textbf{0.63} & 0.42 & 0.38 & 0.18 & 0.29 & 0.15 \\
CBI         & \textbf{0.71} & 0.64 & 0.58 & 0.47 & 0.54 & 0.43 \\
ReefSet     & \textbf{0.44} & 0.41 & --   & 0.28 & 0.32 & 0.24 \\
EFB         & \textbf{0.57} & 0.50 & 0.45 & 0.35 & 0.40 & 0.29 \\
HumBugDB    & \textbf{0.42} & 0.38 & 0.35 & 0.27 & 0.30 & 0.23 \\
Dogs        & \textbf{0.79} & 0.71 & 0.64 & 0.51 & 0.56 & 0.45 \\
\bottomrule
\end{tabular}
\end{table*}

\begin{table*}[t]
\centering
\caption{\textbf{Needle counts across datasets and embedding models.} Lower is better.}
\label{tab:appendix_embedding_needles}
\tiny
\setlength{\tabcolsep}{5pt}
\begin{tabular}{lcccccc}
\toprule
\textbf{Dataset} & \textbf{Perch 2.0} & \textbf{Perch 1.0} & \textbf{SurfPerch} & \textbf{BirdNET} & \textbf{BEATS} & \textbf{AudioMAE} \\
\midrule
Watkins     & \textbf{47}   & 50   & 55   & 59   & 61   & 65 \\
CBI         & \textbf{11}   & 13   & 15   & 18   & 16   & 19 \\
ReefSet     & \textbf{1357} & 1451 & --   & 1644 & 1604 & 1714 \\
EFB         & \textbf{781}  & 859  & 902  & 961  & 918  & 997 \\
HumBugDB    & \textbf{457}  & 497  & 526  & 565  & 546  & 592 \\
Dogs        & \textbf{3}    & 4    & 5    & 6    & 5    & 7 \\
\bottomrule
\end{tabular}
\end{table*}

\begin{table*}[t]
\centering
\caption{\textbf{Fraction of rare samples surfaced during active labeling.} Higher is better. Values are reported after the same fixed annotation budget with margin sampling.}
\label{tab:appendix_embedding_rare}
\tiny
\setlength{\tabcolsep}{5pt}
\begin{tabular}{lcccccc}
\toprule
\textbf{Dataset} & \textbf{Perch 2.0} & \textbf{Perch 1.0} & \textbf{SurfPerch} & \textbf{BirdNET} & \textbf{BEATS} & \textbf{AudioMAE} \\
\midrule
Watkins     & \textbf{0.62} & 0.54 & 0.46 & 0.37 & 0.41 & 0.31 \\
CBI         & \textbf{0.81} & 0.72 & 0.66 & 0.55 & 0.60 & 0.49 \\
ReefSet     & \textbf{0.40} & 0.36 & --   & 0.25 & 0.28 & 0.21 \\
EFB         & \textbf{0.69} & 0.60 & 0.54 & 0.41 & 0.47 & 0.36 \\
HumBugDB    & \textbf{0.49} & 0.43 & 0.39 & 0.31 & 0.35 & 0.27 \\
Dogs        & \textbf{0.91} & 0.84 & 0.76 & 0.63 & 0.70 & 0.58 \\
\bottomrule
\end{tabular}
\end{table*}

Across bioacoustic datasets, Perch 2.0 produces the strongest cluster alignment, the fewest needles, and the earliest rare-sample recovery, while weaker or less specialized encoders leave more minority examples geometrically obscured. This pattern is mild on easier datasets such as Dogs and CBI, but becomes much more consequential on harder discovery-limited settings such as ReefSet and the camera-trap datasets. These tables therefore support the claim that improving representation geometry can matter more for transductive active labeling than switching between acquisition heuristics alone.

\newpage
\clearpage

\subsection{Additional Figures}

\begin{figure}[h]
    \centering
    \includegraphics[width=\linewidth]{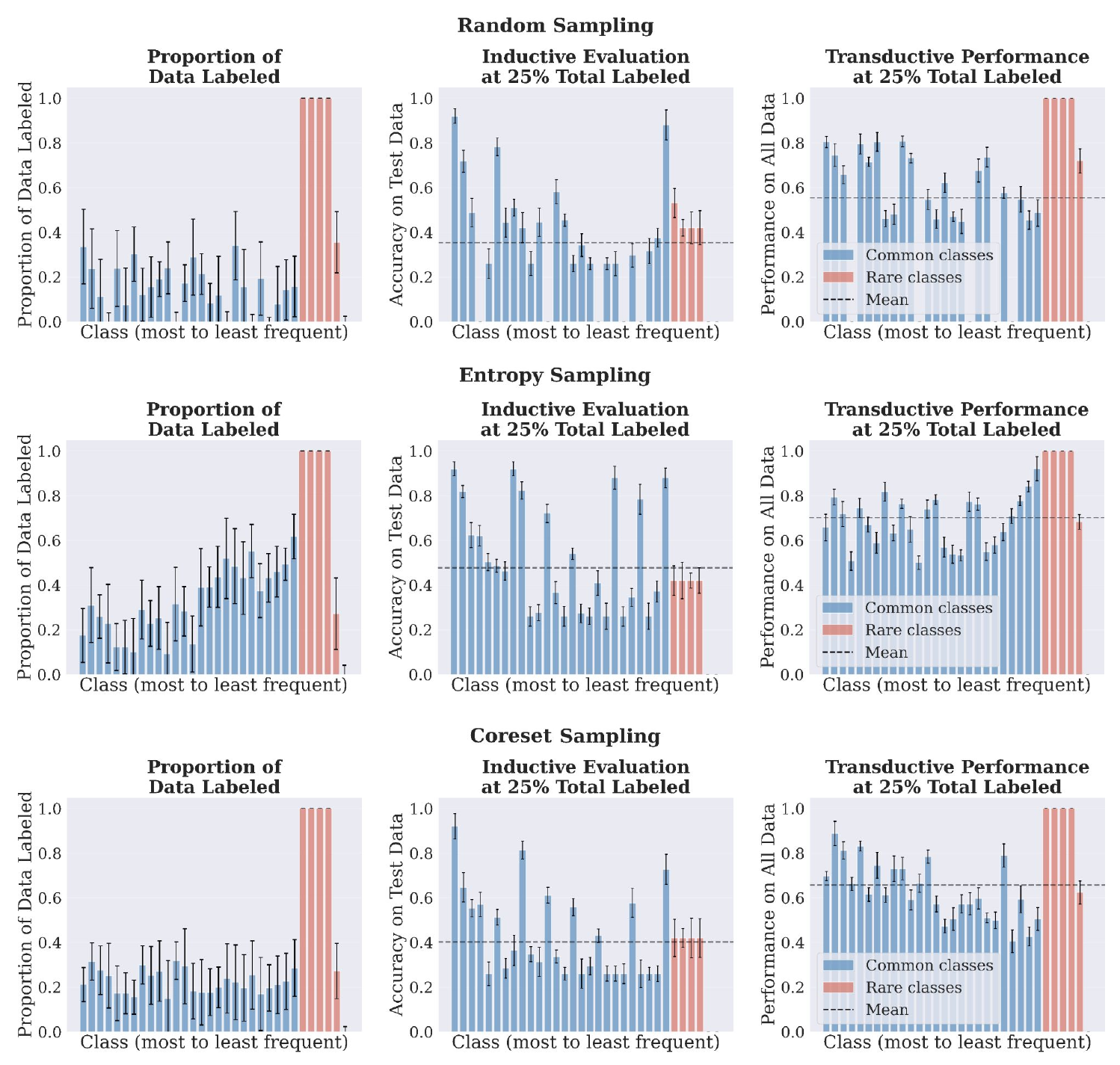}
    \caption{\textbf{Rare classes are sampling-limited, across acquisition strategies.} Results are shown after labeling 25\% of the dataset over the Watkins Marine Mammal Dataset, across random, entropy, and coreset strategies, across five seeds.}
    \label{fig:watkins_strat}
\end{figure}

\begin{figure}[h]
    \centering
    \includegraphics[width=0.95\linewidth]{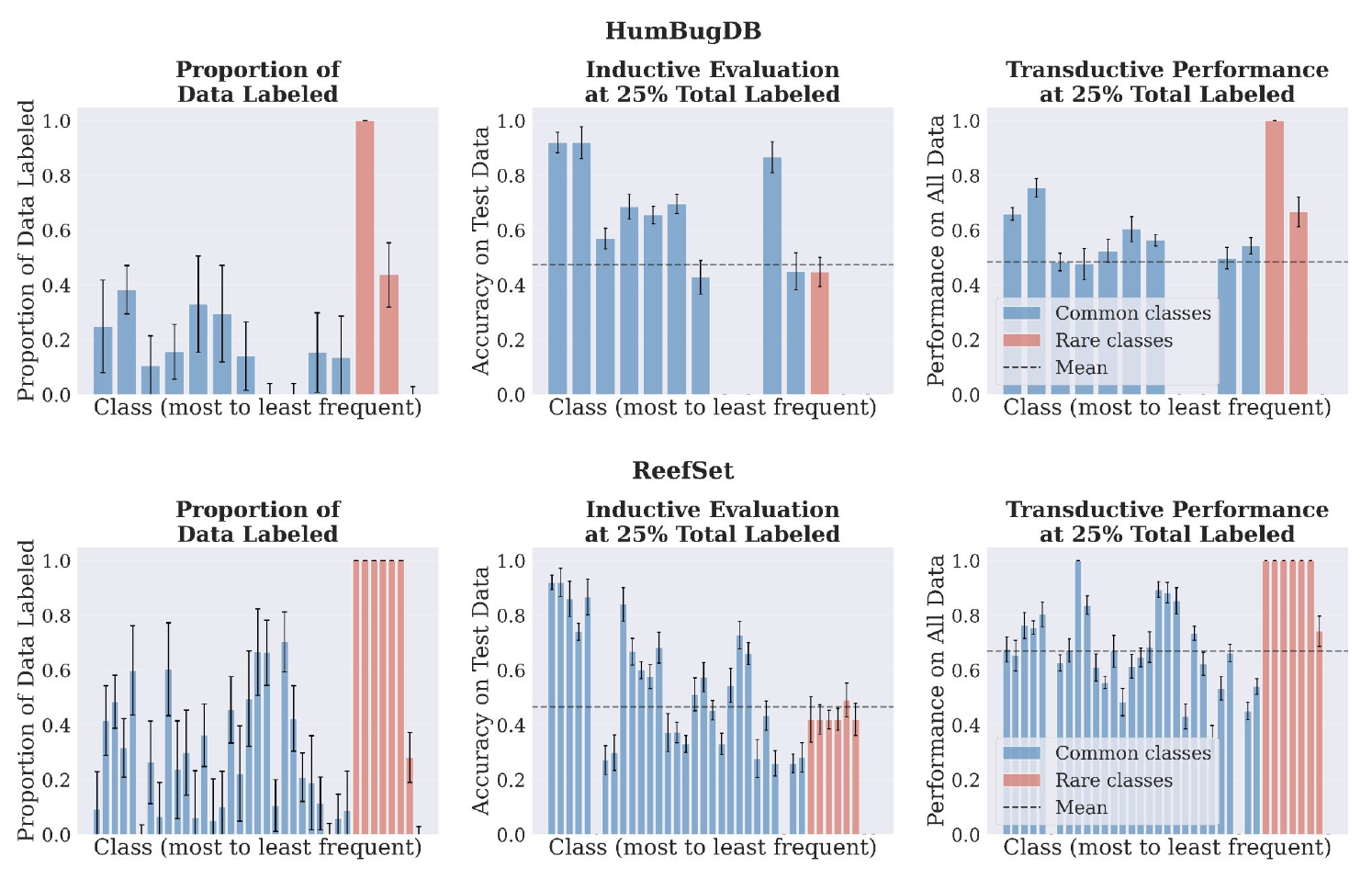}
    \caption{\textbf{Rare classes are sampling-limited, across bioacoustic datasets}. Results are shown after labeling 25\% per dataset, with margin sampling across five seeds.}
    \label{fig:datasets_sampling_1}
\end{figure}

\begin{figure}[h]
    \centering
    \includegraphics[width=0.95\linewidth]{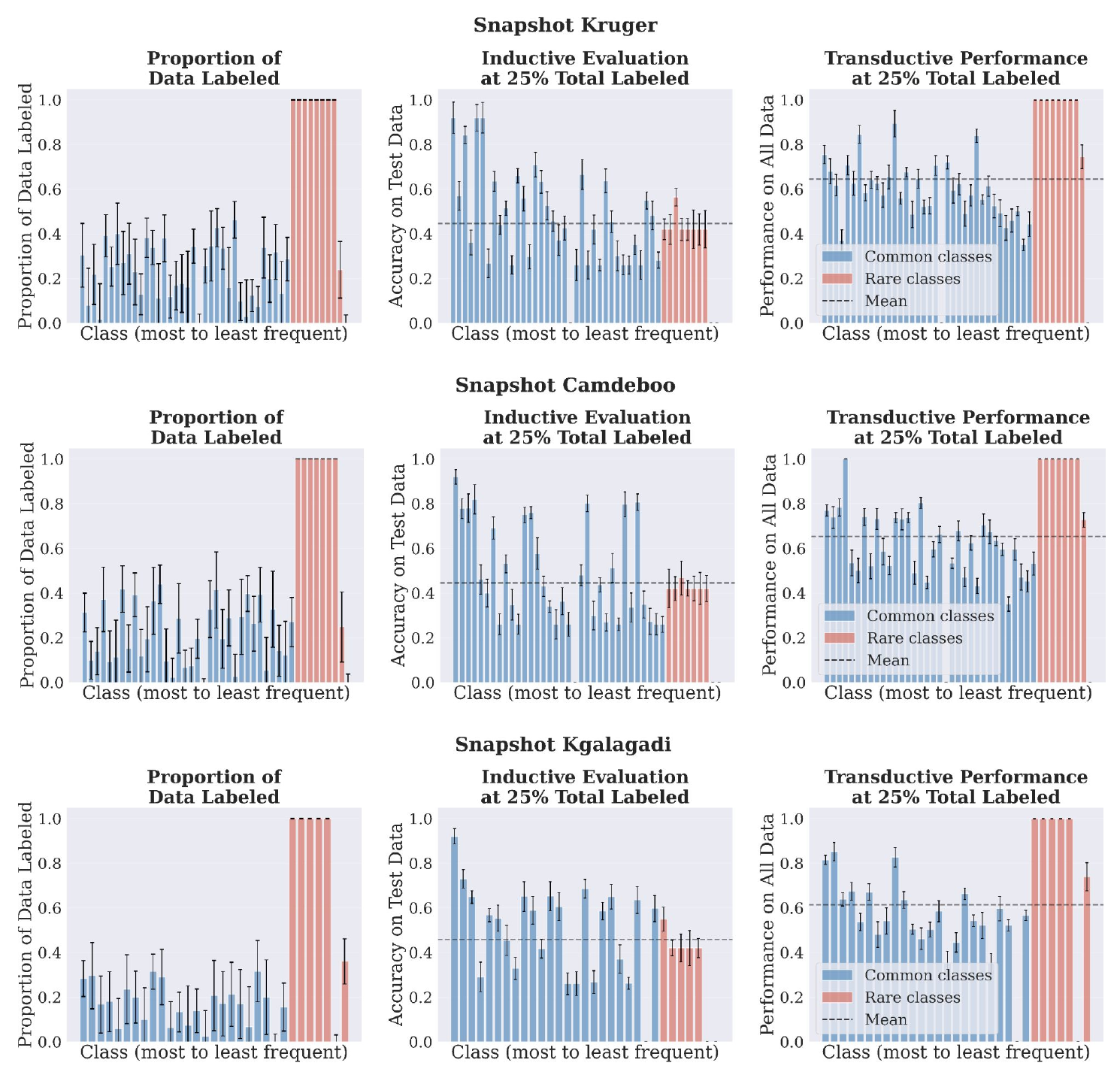}
    \caption{\textbf{Rare classes are sampling-limited, across image datasets}. Results are shown after labeling 25\% per dataset, with margin sampling across five seeds.}
    \label{fig:datasets_sampling_2}
\end{figure}

\begin{figure}[h]
    \centering
    \includegraphics[width=0.95\linewidth]{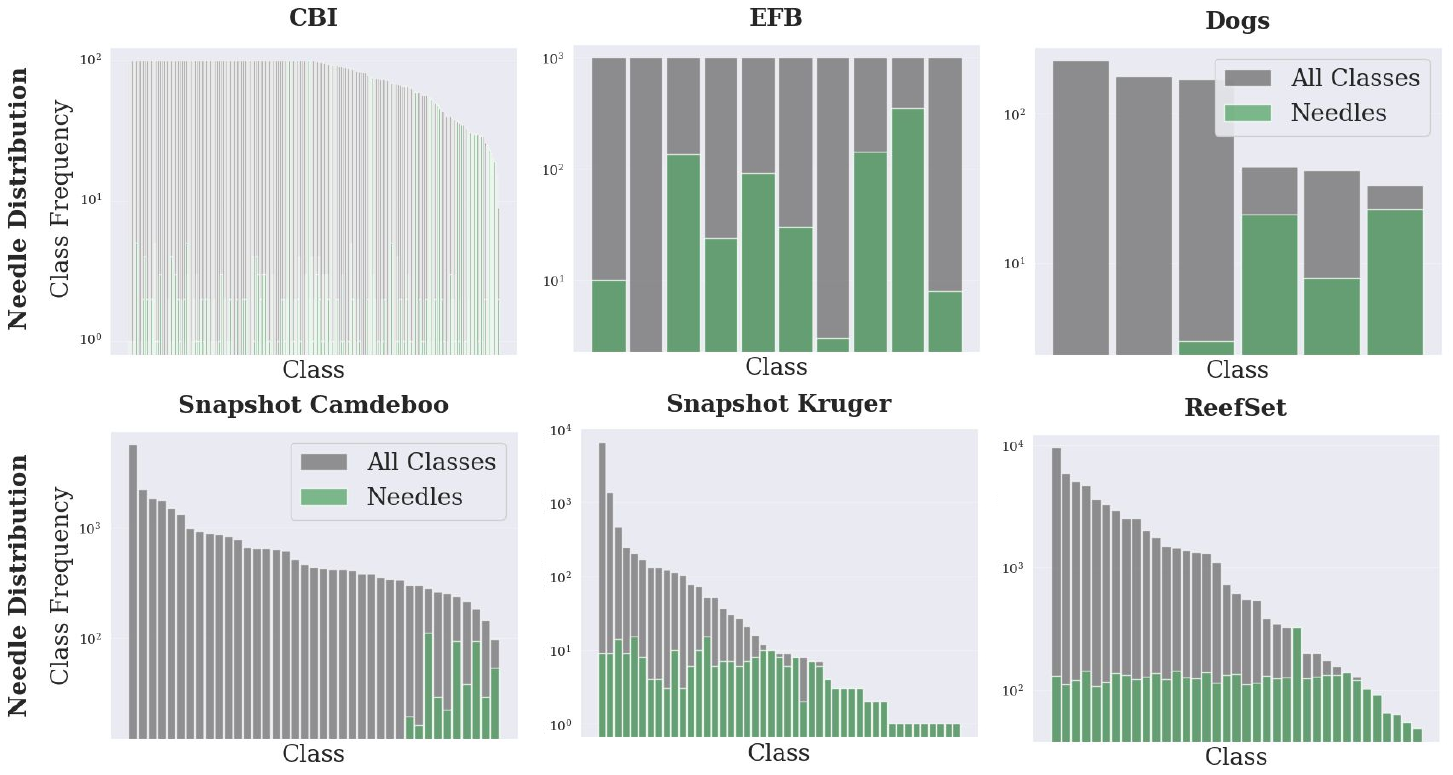}
    \caption{\textbf{Needles concentrate in the long tail, across datasets.} Across image and bioacoustic datasets, needles are both skewed toward minority classes and systematically harder to sample due to dense, mixed local neighborhoods in embedding space. EFB is an exception, where the data are uniformly distributed across the classes.}
    \label{fig:needle_dist_ext}
\end{figure}

\begin{figure}[h]
    \centering
    \includegraphics[width=\linewidth]{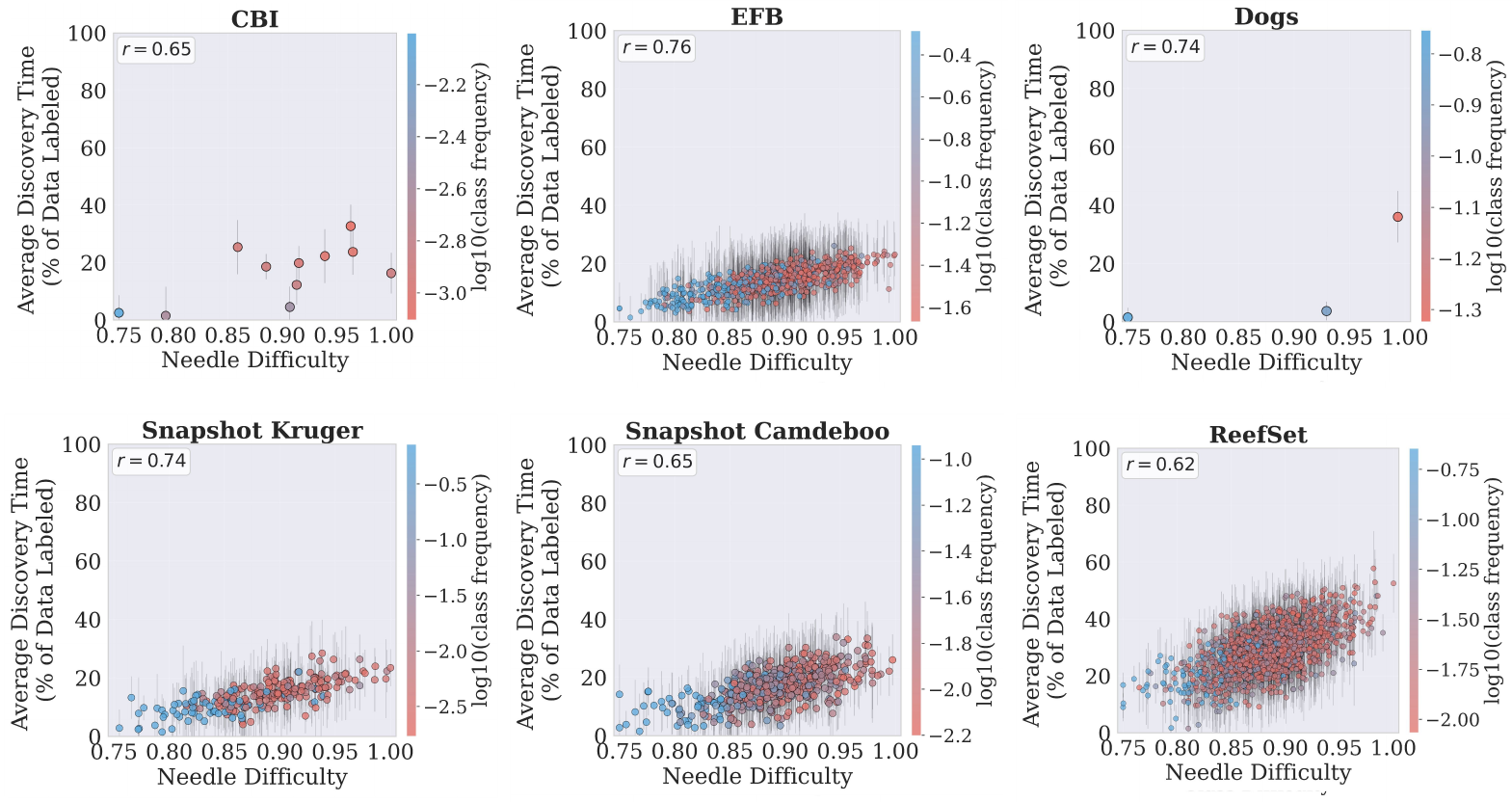}
    \caption{\textbf{Across datasets, difficult classes are discovered later.} We compare the average difficulty for all data in a class against the cycle at which the class is first discovered, with error bars reported across all baseline sampling strategies.}
    \label{fig:discovery_other}
\end{figure}

\begin{figure}[h]
    \centering
    \includegraphics[width=0.5\linewidth]{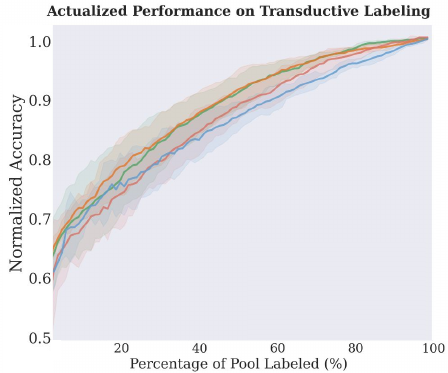}
    \caption{\textbf{Effect of batch size under transductive labeling.} Normalized accuracy under margin sampling (seed 42), averaged across all six datasets, as a function of the percentage of the pool labeled for different batch sizes (labels acquired per iteration). Batch size has minimal effect on transductive labeling performance because as the batch size increases and more data are labeled per iteration, there are more data points automatically considered ``correct.'' Smaller batch sizes may achieve slightly higher efficiency due to more frequent model updates.}
    \label{fig:budget}
\end{figure}

\clearpage
\newpage
\subsection{Sensitivity to Stopping Criteria Parameters}
\label{app:sensitivity}

For each of the following sensitivity analyses over the parameter-tuning datasets in our stopping criteria analysis, each panel varies one stopping parameter while holding the others fixed at their default values. Blue and red curves show common-class and rare-class transductive labeling performance, respectively, and the green dashed curve shows the percentage of data labeled. Shaded regions indicate variability across simulated repeats. The vertical dashed line marks the chosen default value, and the dotted line marks the best-performing value for this dataset under a balanced accuracy-versus-labeling-cost criterion. The stopping behavior is reasonably stable across a broad range of settings, with predictable trade-offs between performance and annotation budget.

\begin{figure}[h]
    \centering
    \includegraphics[width=\linewidth]{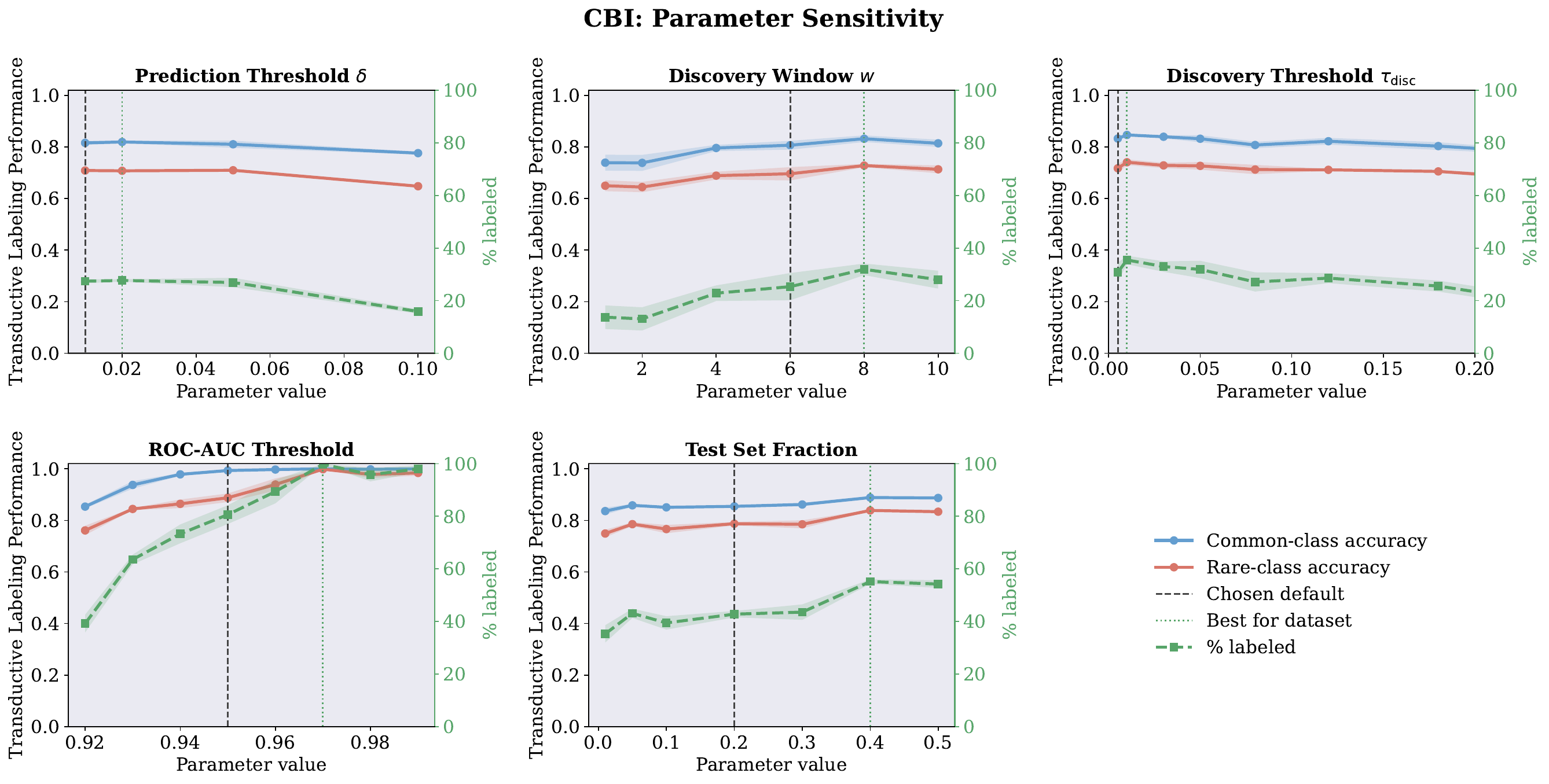}
    \caption{\textbf{Sensitivity to Stopping Criteria Parameters on the CBI Dataset.} }
    \label{fig:stopping_cbi}
\end{figure}

\begin{figure}[h]
    \centering
    \includegraphics[width=\linewidth]{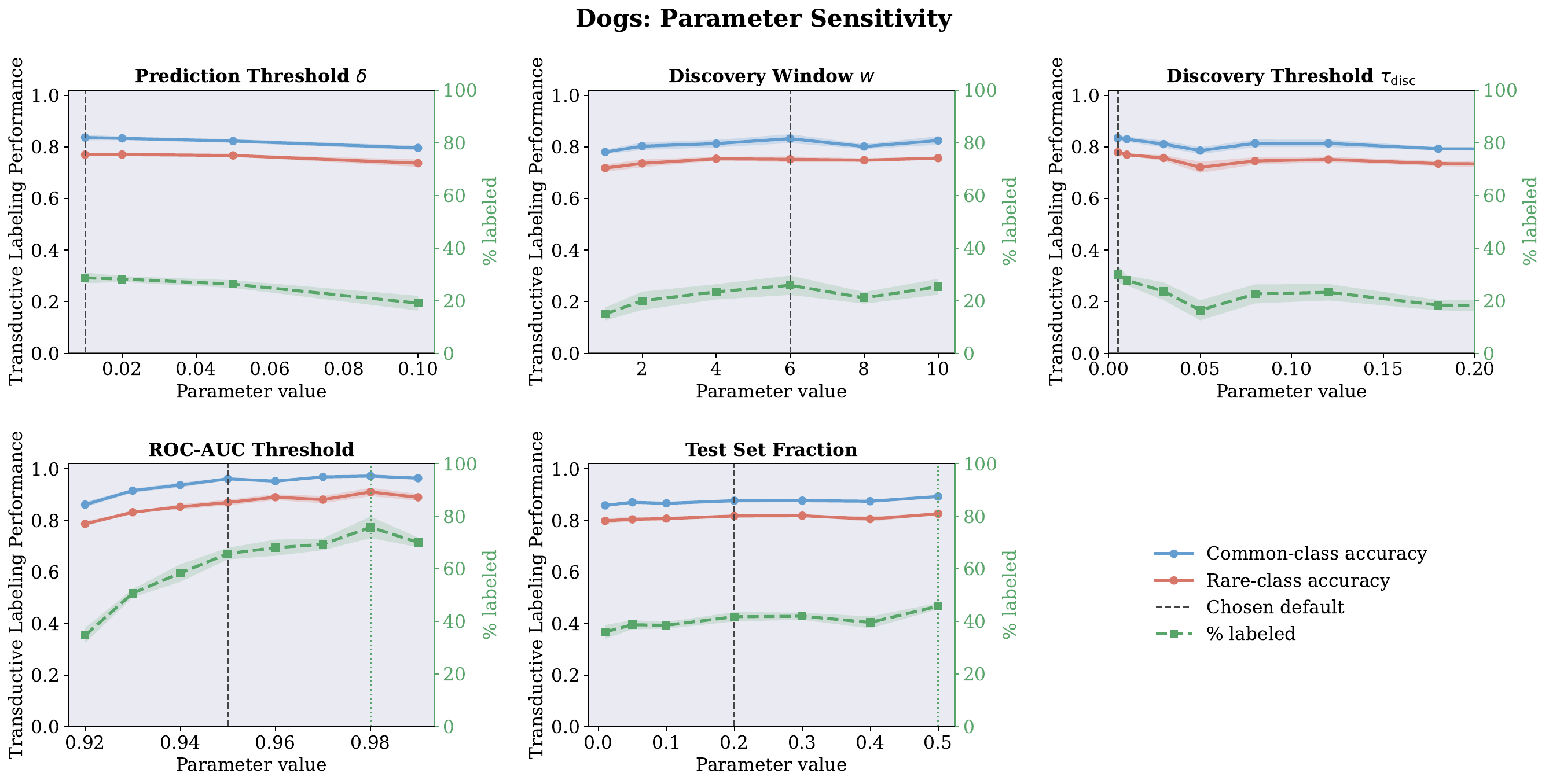}
    \caption{\textbf{Sensitivity to Stopping Criteria Parameters on the Dogs Dataset.} }
    \label{fig:stopping_dogs}
\end{figure}

\begin{figure}[h]
    \centering
    \includegraphics[width=\linewidth]{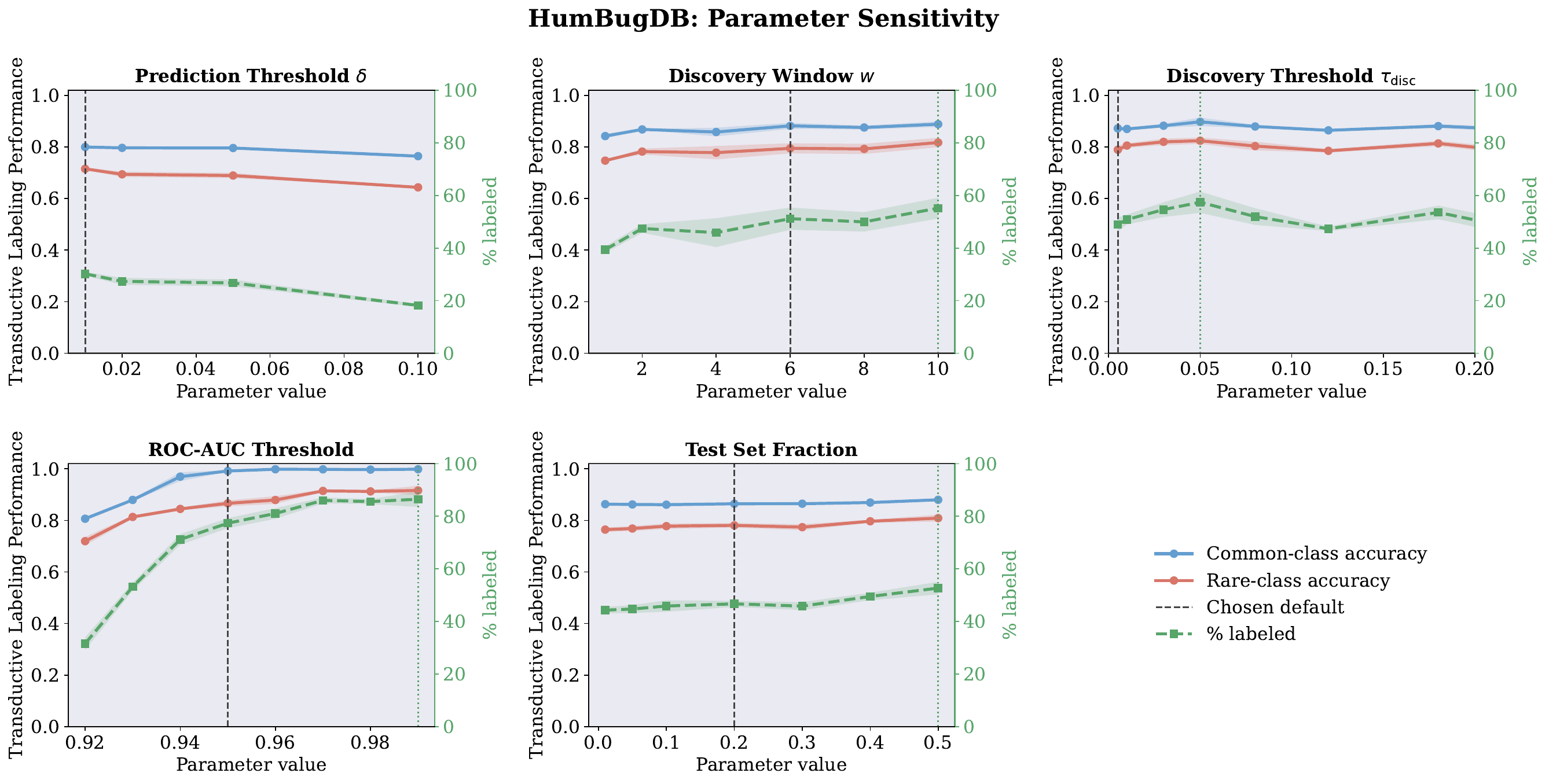}
    \caption{\textbf{Sensitivity to Stopping Criteria Parameters on the HumBugDB Dataset.} }
    \label{fig:stopping_humbug}
\end{figure}

\begin{figure}[h]
    \centering
    \includegraphics[width=\linewidth]{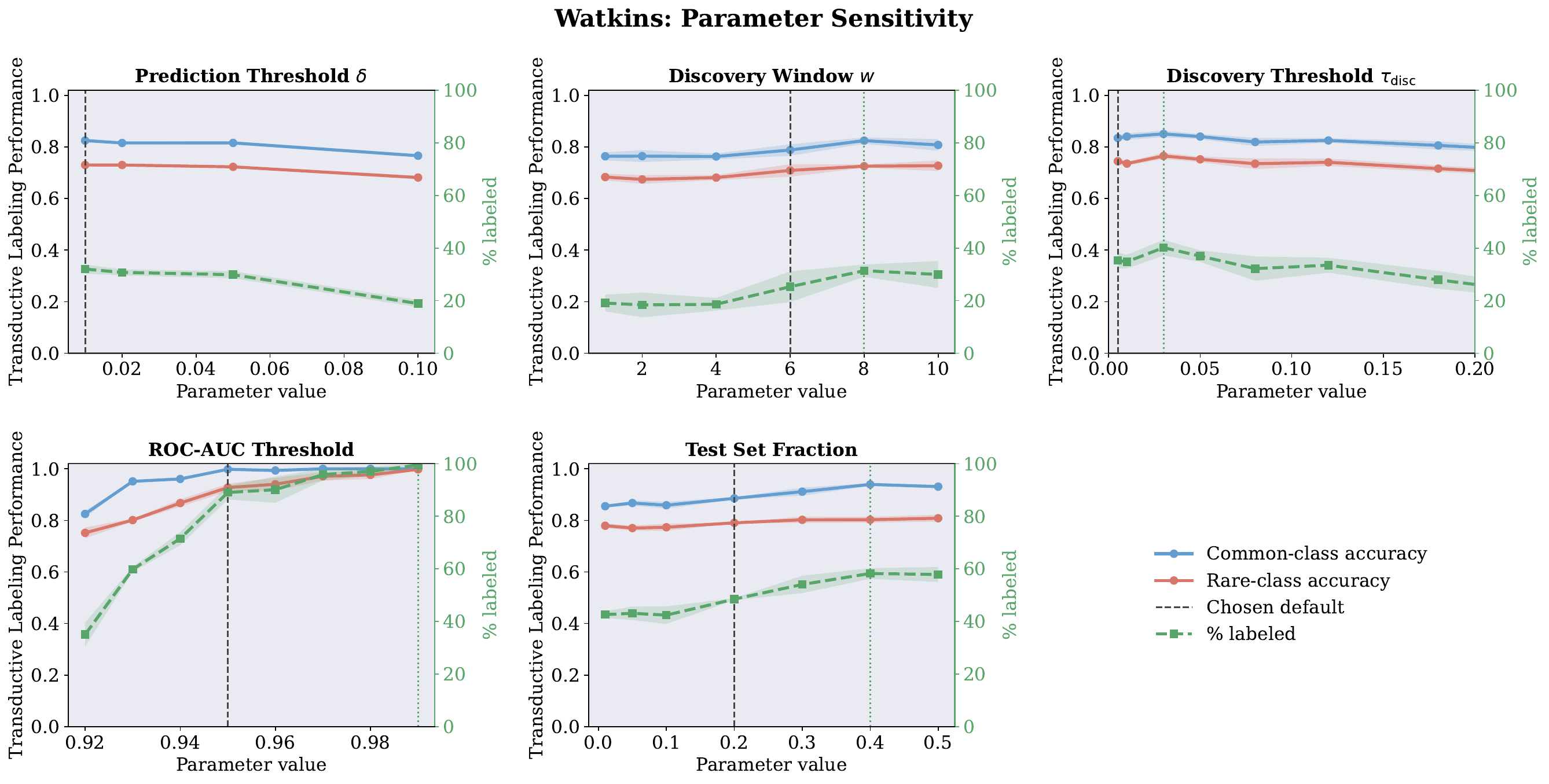}
    \caption{\textbf{Sensitivity to Stopping Criteria Parameters on the Watkins Dataset.} }
    \label{fig:stopping_watkins}
\end{figure}

\begin{figure}[h]
    \centering
    \includegraphics[width=\linewidth]{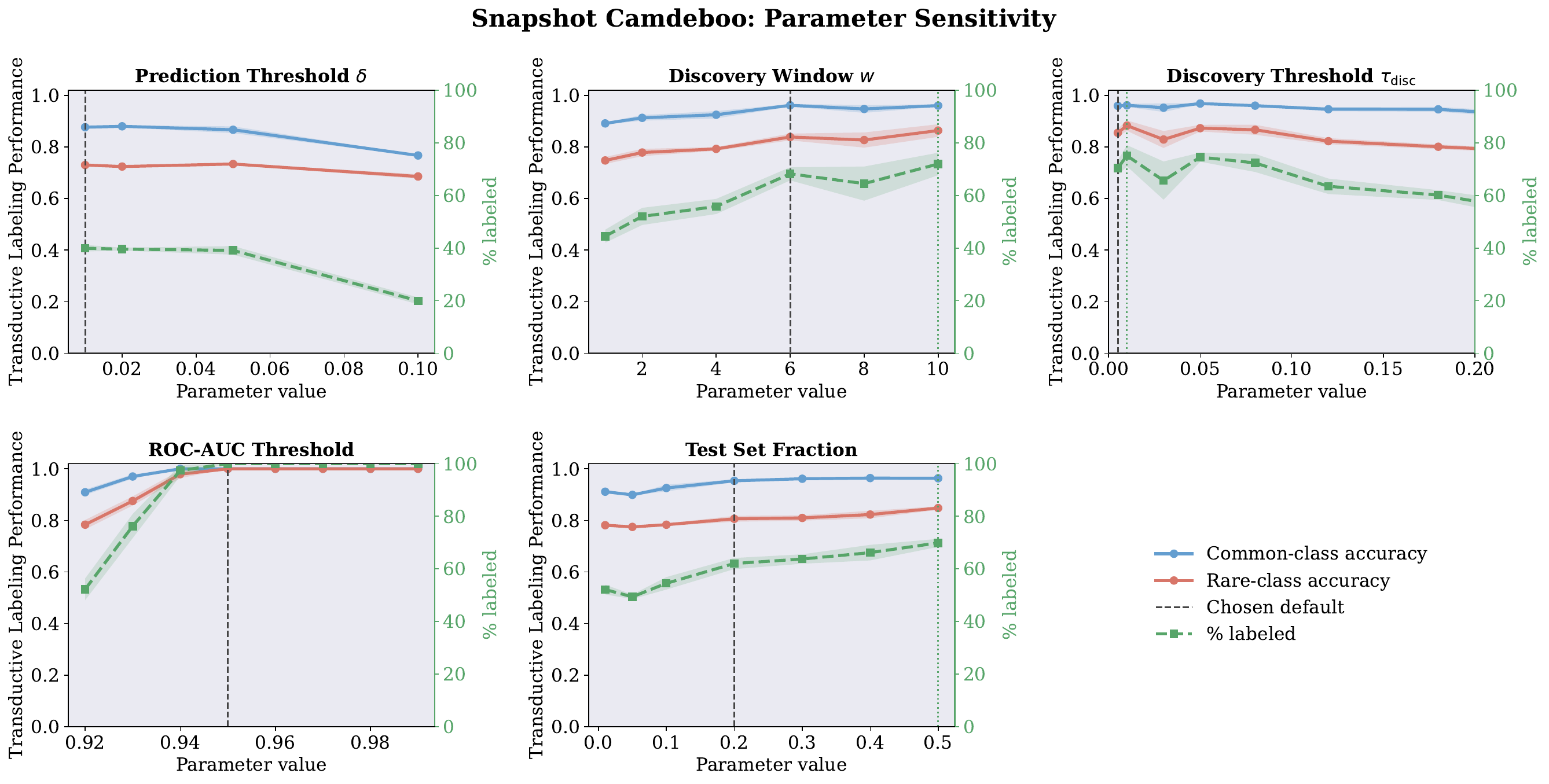}
    \caption{\textbf{Sensitivity to Stopping Criteria Parameters on the Snapshot Camdeboo Dataset.} }
    \label{fig:stopping_cdb}
\end{figure}